%% file: arxiv.tex
\documentclass{article} 
\usepackage{iclr2027_conference,times}

\input{math_commands.tex}

\usepackage{hyperref}
\usepackage{url}

\usepackage{booktabs}
\usepackage{enumitem}
\usepackage{graphicx}
\usepackage{subcaption}
\usepackage{multirow}
\usepackage{wrapfig}

\usepackage{longtable}

\input{sections/macros}

\title{\KairosEnd: A Dataset for Fine-Grained Video-Language Modeling over Space, Time, and Dynamics}

\author{%
\textbf{
Ruibo Ming\textsuperscript{1},
Lei Sun\textsuperscript{1},
Deheng Zhang\textsuperscript{1},
He Zhang\textsuperscript{2},
Jialu Li\textsuperscript{2},
Jian Wang\textsuperscript{3},
Zhendong Li\textsuperscript{1},}\\
\textbf{
Mengshun Hu\textsuperscript{1},
Danda Pani Paudel\textsuperscript{1},
Luc van Gool\textsuperscript{1},
Jinjin Gu\textsuperscript{1}}\\
\textsuperscript{1}INSAIT, Sofia University ``St. Kliment Ohridski'' \quad
\textsuperscript{2}Adobe Research\quad
\textsuperscript{3}Snap Research \\
}

\iclrfinalcopy 
\begin{document}

\maketitle

\fancyhead{}
\renewcommand{\headrulewidth}{0pt}

\input{fig_table/teaser_image}

\begin{abstract}
\input{sections/abstract}
\end{abstract}

\input{sections/introduction}

\input{sections/dataset}

\input{sections/benchmark}

\input{sections/conclusion}

\bibliography{iclr2027_conference}
\bibliographystyle{iclr2027_conference}

\appendix
\section*{Appendix}

\input{sections/relatedworks}

\input{sections/appendix}


\end{document}

%% file: math_commands.tex
\usepackage{amsmath,amsfonts,bm}

\def\eqref#1{equation~\ref{#1}}

\def\1{\bm{1}}

\DeclareMathAlphabet{\mathsfit}{\encodingdefault}{\sfdefault}{m}{sl}
\SetMathAlphabet{\mathsfit}{bold}{\encodingdefault}{\sfdefault}{bx}{n}



%% file: sections/macros.tex
\newcommand{\Kairos}{\textsc{Kairos}~}
\newcommand{\KairosEnd}{\textsc{Kairos}}
\newcommand{\KairosBench}{\textsc{Kairos}-Bench~}
\newcommand{\KairosBenchEnd}{\textsc{Kairos}-Bench}

\newcommand{\videoDomains}{12~}
\newcommand{\videoCategories}{35~}
\newcommand{\videoScenarios}{202~}
\newcommand{\videoURLs}{28,282~}
\newcommand{\videoYouTubeURLs}{15,054~}
\newcommand{\videoBilibiliURLs}{13,228~}

\newcommand{\dataVideos}{19,004~}
\newcommand{\dataYouTubeVideos}{9,434~}
\newcommand{\dataBilibiliVideos}{9,570~}
\newcommand{\dataHours}{5,420~}
\newcommand{\dataGPUHours}{1,355~}  
\newcommand{\dataMinMinutes}{10~}
\newcommand{\dataMeanMinutes}{17~}
\newcommand{\dataMaxMinutes}{30~}
\newcommand{\dataMaxMinutesEnd}{30}

\newcommand{\benchCapabilities}{17~}
\newcommand{\benchSourceTypes}{10~}
\newcommand{\benchVideos}{820~}
\newcommand{\benchRawMCQs}{25,707~}
\newcommand{\benchAuditedMCQs}{6,277~}
\newcommand{\benchReviewedMCQs}{2,870~}

\newcommand{\evalModels}{21~}
\newcommand{\evalClosedModels}{7~}
\newcommand{\evalOpenModels}{14~}

\newcommand{\trainQuestions}{232,101~}
\newcommand{\trainEpochs}{1~}
\newcommand{\trainFrames}{32~}

%% file: fig_table/teaser_image.tex
\begin{center}
    \vspace{-8mm}    
    \includegraphics[width=\textwidth]{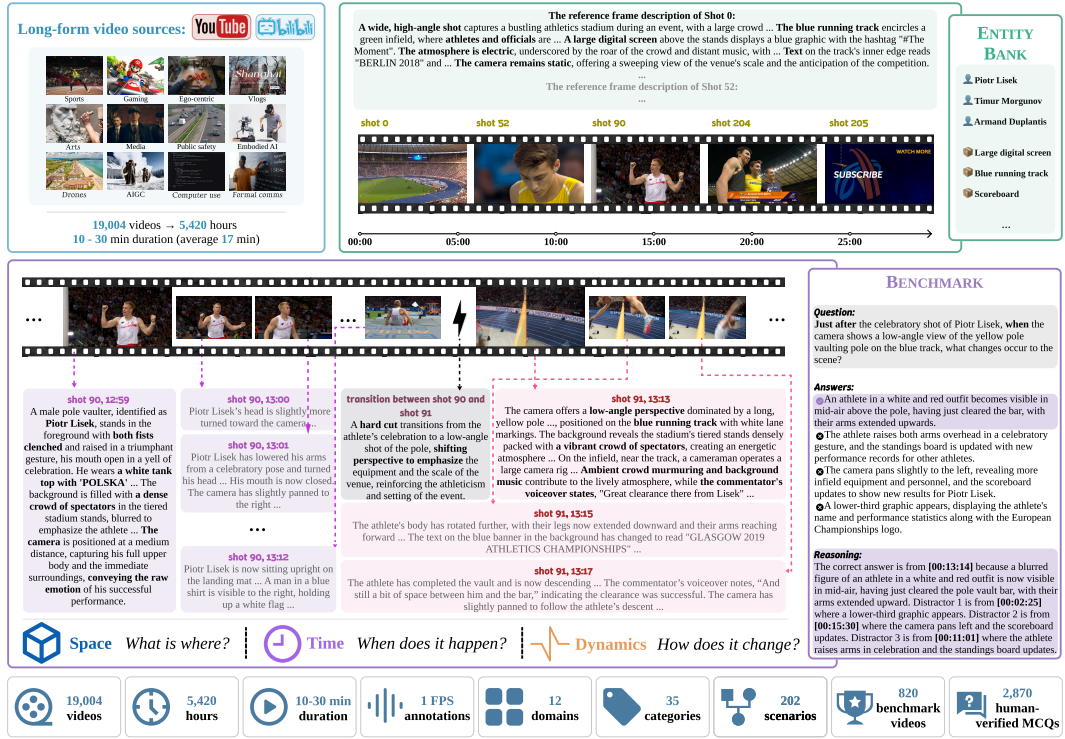}
    \captionof{figure}{\Kairos represents long-form videos as structured, time-resolved annotation streams. Each video is decomposed into shots, with each shot annotated by a reference-frame description, an in-shot differential chain, and a video-level entity bank that links recurring identities across shots. At 1 FPS, the annotations capture three axes of fine-grained video understanding: \emph{Space}, \emph{Time}, and \emph{Dynamics}. \KairosBench is derived from \Kairos by converting these structured annotations into temporally grounded questions, whose answers are tied to explicit evidence spans.}
    \label{fig:teaser}
\end{center}

%% file: sections/abstract.tex
Many emerging video language modeling tasks require systems to move beyond clip-level abstraction and model visual content as it unfolds over extended time horizons. However, most existing video datasets rely on coarse or sparsely aligned supervision, which compresses temporal variation and limits the ability of models to learn reusable representations of continuous visual dynamics. We introduce \KairosEnd, a video dataset for video-language modeling with time-resolved annotations. \Kairos consists of long-duration videos, ranging from ten minutes to half an hour, annotated with fine-grained temporal alignment. The annotations capture ongoing actions, entity appearances and attributes, interactions, and evolving contextual cues along the video timeline. This time-resolved structure supports fine-grained evaluation, long-range modeling and reasoning, instruction data construction, representation learning, and video generation. \Kairos provides a general-purpose foundation for modeling visual experiences over time.

%% file: sections/introduction.tex
\section{Introduction}
Video-language modeling~\citep{venugopalan2015sequence,sun2019videobert,zhu2020actbert,li2020hero,luo2020univl,fu2021violet,li2022align,xu2021videoclip,alayrac2022flamingo,li2023blip,zhang2023video,li2024llava} seeks to represent and reason over video content through language as it unfolds over both space and time.
In a video, entities appear, interact, and evolve continuously, forming structured patterns across both spatial layouts and temporal progressions.
Such spatiotemporal evolution gives rise to rich dynamics, including event progression, state transitions, and causal dependencies, and provides the foundation for grounded reasoning about what happens, why it happens, and what may follow.
Therefore, fine-grained video-language modeling requires jointly modeling Space, Time, and Dynamics:
\textbf{Space} specifies what is present and how it is spatially arranged,
\textbf{Time} captures when events occur across the video timeline, and
\textbf{Dynamics} characterizes how entities, actions, and contexts evolve and interact to form coherent video-level understanding.

A fundamental barrier to achieving this goal lies in the data. Significant efforts have been made to advance video understanding, including construction of video caption datasets~\citep{chen2024sharegpt4video,Farre2024FineVideo}, video instruction tuning and conversational video datasets~\citep{luo2023valley,zhang2024llava,chen2024sharegpt4video,ren2024timechat}, and recent long video benchmarks~\citep{chen2024cgbenchcluegroundedquestionanswering,Qin_2025,mangalam2023egoschema,wu2024longvideobench,cheng2025videoholmesmllmthinklike,fu2025video,chandrasegaran2024hourvideo1hourvideolanguageunderstanding,li2024mvbench,wang2025lvbench}. Despite these advances, almost all prior works still fall short on all three fronts outlined above.
\begin{enumerate}[leftmargin=*, labelsep=0.4em, itemsep=0pt, topsep=0pt, partopsep=0pt, parsep=0pt]
    \item \textbf{Space:} Existing datasets still lack spatial granularity. Most rely on holistic video captions and place far less emphasis on fine-grained visual annotation than image datasets. As a result, their annotations typically foreground only the most salient objects, leaving many entities, attributes, and relations essential for a comprehensive account of video content unspecified.
    \item \textbf{Time:} Existing datasets offer limited temporal resolution. Most provide annotations at the clip level, such as a single caption or a set of questions, without fine-grained temporal alignment. While such annotations capture high-level semantics, they fail to reflect how visual content, events, and scene states evolve over time.
    \item \textbf{Dynamics:}
    Existing datasets, especially benchmarks, largely fail to probe fine-grained spatiotemporal grounding.
    This includes aspects such as pinpointing precisely when an event occurs, tracing how states change over time, and examining how observations from different moments jointly support reasoning about temporal order, interactions, and causality.
    As a result, they provide only limited supervision and evaluation of such dynamics, leaving models inadequately assessed in their ability to perform grounded reasoning over dynamic video structure.
\end{enumerate}
These limitations are especially pronounced in long-form videos (e.g., longer than 10 minutes), where richer temporal dependencies accumulate over extended durations but remain sparsely annotated.

We propose \KairosEnd, a new dataset for fine-grained video-language modeling over Space, Time, and Dynamics.
The core of \Kairos lies an automated annotation pipeline that produces fine-grained spatiotemporal descriptions at multiple levels of granularity: within each shot, it records detailed visual content; across frames within a shot, it captures temporal variation and scene evolution; and over the full video, it integrates these observations into spatially detailed and temporally resolved descriptions.
A key component of this framework is a unified, text-centric representation that maintains consistency across long video contexts.
\Kairos integrates entity consistency directly into frame-level annotations, allowing recurring entities and semantic attributes to be resolved over time.
This design links entities across distant segments through language, while incorporating multimodal signals such as speech and environmental audio into a coherent narrative.

As demonstrated in \figurename~\ref{fig:teaser}, built on this pipeline, \Kairos provides rich, time-resolved annotations for constructing both training data and benchmarks for video-language models.
Since evidence may be distributed across distant temporal segments, \Kairos supports tasks such as cross-temporal reference resolution, causal reasoning, and compositional reasoning.
Its annotations can be converted into dense dynamic captions, question-answer pairs, and explicit reasoning paths grounded in temporally localized evidence.
This enables questions about what happened, when an entity appeared, how entities interacted, and how a scene evolved over time, together with rationales that connect answers to the corresponding video evidence.

Our final dataset contains \dataVideos videos with an average duration of \dataMeanMinutes minutes, totaling \dataHours hours of annotated video across \videoScenarios diverse scenario types.
We focus on videos ranging from \dataMinMinutes to \dataMaxMinutes minutes in length to better reflect real-world temporal horizons while maintaining high annotation quality.
This regime is particularly important because the limitations of existing datasets become even more severe in long-form videos, where richer temporal dependencies accumulate over extended durations while annotations remain sparse.
From a subset of \benchVideos videos, we further construct a benchmark of \benchReviewedMCQs questions, all of which are passed through sanity checks and human verification.
These questions are temporally grounded to specific moments in the video and include explicit reasoning rationales.
As illustrated in \figurename~\ref{fig:teaser}, \Kairos exhibits a substantial leap compared to caption-based datasets in annotation density and information richness.

Our extensive experiments on state-of-the-art video language models~\citep{openai2026gpt55,geminiteam2026gemini31pro,Intelligence2024,bytedanceseed2026seed2,chen2024internvl,bai2025qwen3,coreteam2025mimounlockingreasoningpotential} reveal that performance degrades when the evidence spans several shots or the whole video. Besides, we highlight that fine-tuning on training data derived from \Kairos can improve the performance of video-language models on other long video benchmarks.

Our contributions are threefold.
\begin{enumerate}[leftmargin=*, labelsep=0.4em]
    \item We introduce \KairosEnd, a new long-form video-language dataset designed for fine-grained modeling of Space, Time, and Dynamics.

    \item We develop an automated annotation pipeline that produces multi-level spatiotemporal representations for long videos, which captures detailed and structured visual content and enabling coherent tracking of recurring entities and semantic attributes over time.

    \item We construct a temporally grounded benchmark and demonstrate the utility of \Kairos for evaluating and improving video-language models.
\end{enumerate}

%% file: sections/dataset.tex
\section{The \Kairos Dataset}

\noindent\textbf{Annotation format.}
\Kairos represents each video as a dense, temporally unfolding annotation stream.
Rather than assigning a single coarse clip-level description, it refreshes annotations at 1 FPS to capture fine-grained visual, auditory, and semantic changes as they occur.
The format also preserves video structure: frames are organized into locally coherent shots, while a video-level entity matching mechanism tracks identities, attributes, and relationships as they persist, disappear, reappear, or are disambiguated across shots.
Formally, each extracted frame is stored as a structured record with four fields: \textbf{(1)} an absolute timestamp, \textbf{(2)} the corresponding video frame, \textbf{(3)} a shot index, and \textbf{(4)} an audio-aware description.
Together, these records provide a temporally grounded interface for fine-grained video understanding, retrieval, temporal reasoning, entity tracking, and video generation.

\vspace{-1mm}
\subsection{\Kairos Annotation Pipeline}
\vspace{-1mm}
The dense and structured nature of \Kairos annotations makes manual annotation prohibitively expensive.
We therefore develop an automated pipeline that first structures the video globally and then performs fine-grained annotation.
As shown in \figurename~\ref{fig:pipeline}, this design preserves both local details and long-range video dynamics under a tractable computational budget.

\noindent\textbf{Video structure parsing.}
We model cross-shot entity consistency to handle entities that reappear across distant shots, undergo visual changes, or are later identified through speech and context, thereby capturing continuity and long-range relationships across the video.
We first employ \texttt{TransNetV2}~\citep{soucek2024transnet} to segment each video into shots.
After shot segmentation, we identify entities appearing in the reference frame of each shot and construct a video-level entity bank.
We define eight types of entities: \texttt{person}, \texttt{animal}, \texttt{object}, \texttt{vehicle}, \texttt{text}, \texttt{location}, \texttt{food}, and \texttt{clothing}.
Each detected entity is associated with a canonical label and a short visual-detail string, describing distinguishing attributes.
We then link each entity mention emitted for a new reference frame to the video-level entity bank. The bank is injected into every reference-frame prompt, which requires that an entity already in the bank be referred to by its canonical name verbatim; a mention whose normalized label exactly matches a bank entry is recorded as a new appearance of that entity. This lets later shots reuse established identities instead of introducing duplicates. In this way, the entity bank serves both as a tracking mechanism and as global context for spatio-temporal descriptions.

\input{fig_table/pipeline_image}

\noindent\textbf{In-shot fine-grained annotation.}
Given the parsed video structure, \Kairos annotates each shot at 1 FPS with an initial-and-subsequent scheme.
The first sampled frame in each shot is treated as the \textit{reference frame} and receives a full moment-level description, including spatial layout, visible entities, attributes, relations, on-screen text, and relevant audio cues.
Subsequent 1 FPS samples are treated as \textit{differential frames}, which record only fine-grained changes relative to the previous second, such as actions, state transitions, motion, interactions, visibility changes, and newly appearing or disappearing entities.
This design reduces redundancy while preserving fine-grained temporal evolution.
The core visual annotation process is driven by \texttt{Qwen3-VL-8B-Instruct}~\citep{bai2025qwen3} and accelerated with \texttt{vLLM}~\citep{kwon2023efficient}.
For each initial or differential frame, the model receives the sampled frame, the current entity bank, and the relevant context history and audio information.
The generated description is structured along three axes: \textit{Space}, describing what is present at a moment; \textit{Time}, anchoring the description to an absolute point on the video timeline; and \textit{Dynamics}, describing how the scene evolves.

\noindent\textbf{Audio and transition annotation.} We combine speech transcription and non-speech audio summarization.
\texttt{faster-whisper-large-v3}~\citep{faster_whisper,radford2023robust} transcribes speech into timestamped sentence-level segments, while \texttt{Qwen2-Audio-7B-Instruct}~\citep{chu2024qwen2} summarizes ambient and event-level sounds in 30-second windows.
Both streams are aligned with the annotation timeline: speech is attached to the corresponding frame, while non-speech summaries are attached only to reference frames to preserve context without redundancy.
\Kairos also explicitly annotates shot-boundary transitions to preserve cross-shot continuity.
For each boundary, the pipeline records the editing technique (hard cut, fade, dissolve, wipe, etc.), the narrative purpose of the cut, and the visual contrast across it.
These descriptions link adjacent shots and prevent the video annotation from becoming a set of disconnected shot-level records.

\noindent\textbf{Annotation computation analysis.} Because \textsc{Kairos} annotations are produced by a fully automated pipeline, the process scales naturally with available GPU parallelism.
In our production setting, one H200 annotates about 4 hours of video per hour; with tensor parallelism over two H200s, the pipeline reaches about $8\times$ real time.
This corresponds to roughly \dataGPUHours H200-hours for the \Kairos dataset of \dataHours hours, with preprocessing and audio analysis running at over $20\times$ real time and contributing little to the overall cost.

\input{fig_table/dataset_categories_image}

\vspace{-1mm}
\subsection{Video Curation}
\vspace{-1mm}
To provide a suitable setting for studying dynamic and fine-grained video understanding, we explicitly prioritize long-form narratives with extended durations and a high density of shot transitions.
Such videos naturally encapsulate rich event progressions, entity interactions, and causal dependencies over time.
We source our raw data from the two most prominent long-form video platforms: \textit{YouTube} and \textit{Bilibili}.
Using an agent-based search strategy, we constructed an initial pool of candidate videos ranging from \dataMinMinutes minutes to 4 hours in length.
This yielded a total of \videoURLs URLs, comprising \videoYouTubeURLs from YouTube and \videoBilibiliURLs from Bilibili.

To ensure semantic diversity, each candidate URL is pre-annotated and filtered using a stringent three-level taxonomy encompassing \videoDomains high-level domains, \videoCategories categories, and \videoScenarios leaf-level scenarios (e.g., \textit{A. Sports $\rightarrow$ A.1. Ball Games $\rightarrow$ A.1.1. Basketball}).
During processing, excessively long videos are segmented into consecutive clips of no more than \dataMaxMinutes minutes to maintain annotation fidelity while preserving long-range context.
Our final annotated dataset consists of \dataVideos high-quality videos, of which \dataYouTubeVideos are from YouTube, and \dataBilibiliVideos are from Bilibili.
The total is \dataHours hours. As illustrated in \figurename~\ref{fig:dataset_categories}, \Kairos exhibits substantial diversity.

%% file: fig_table/pipeline_image.tex
\begin{figure}[t]
\centering
\includegraphics[width=0.8\textwidth]{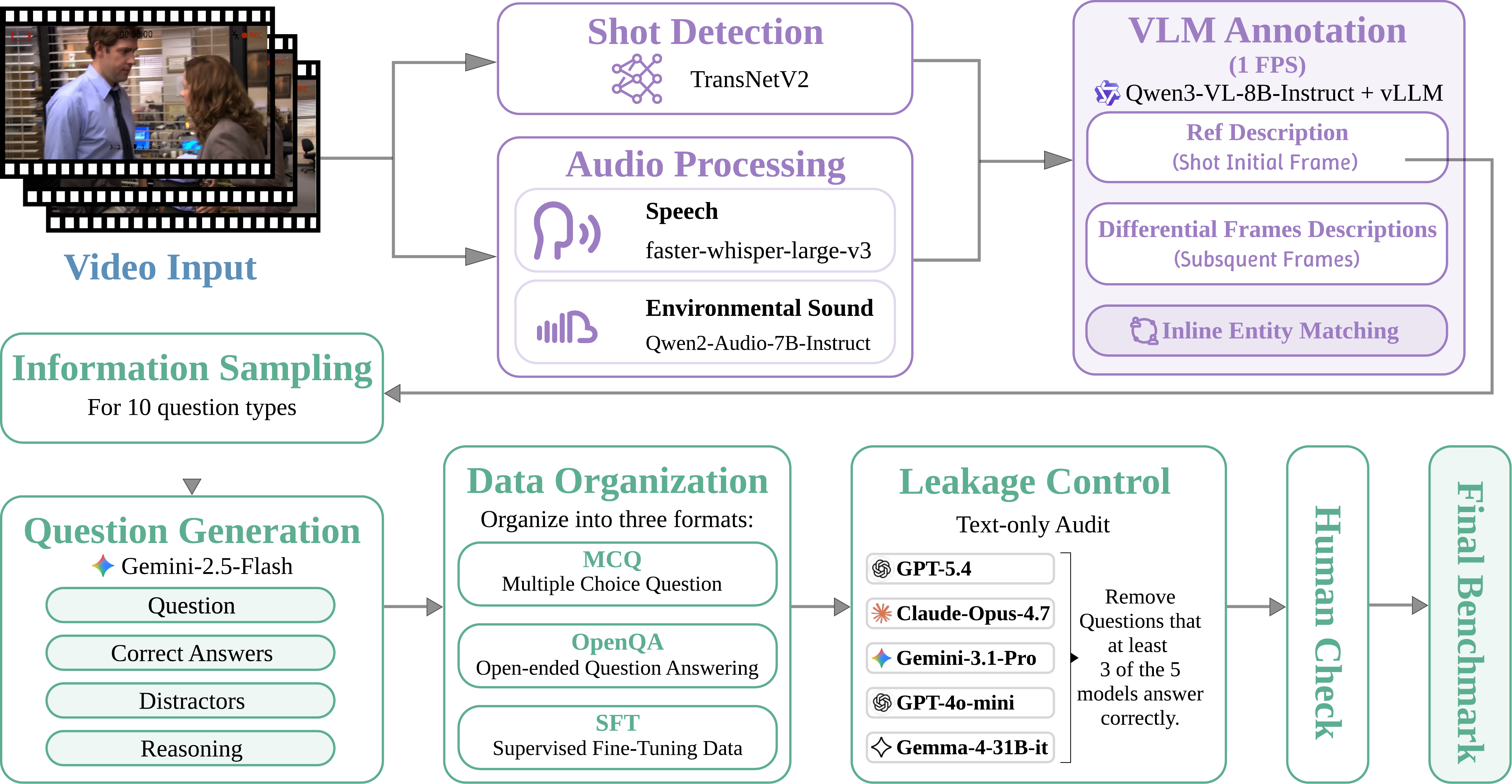}
\caption{\Kairos transforms raw long-form videos into structured, time-resolved annotations that integrate visual, temporal, entity-level, and audio information, supporting dense captioning, reasoning supervision, and \KairosBench construction.}
\vspace{-5mm}
\label{fig:pipeline}
\end{figure}

%% file: fig_table/dataset_categories_image.tex
\begin{figure}[t]
\centering
\includegraphics[width=0.8\textwidth]{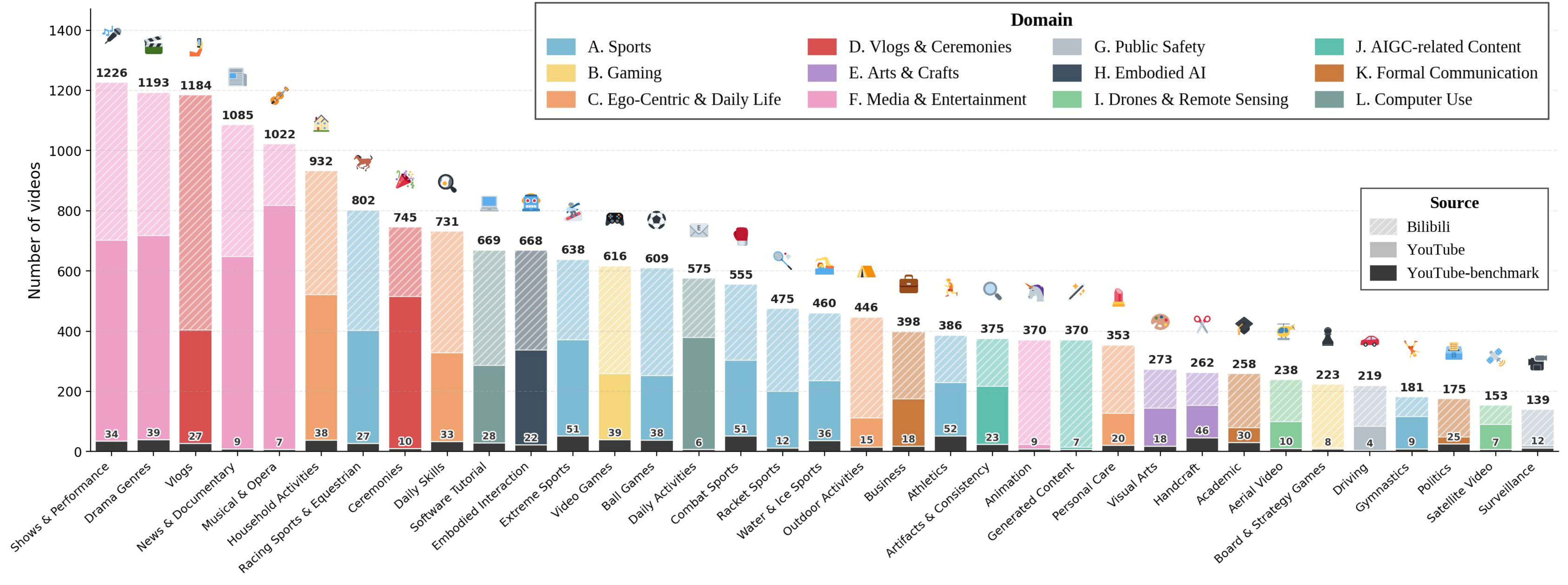}
\vspace{-4mm}
\caption{Per-category video count of \Kairos dataset, grouped by the \videoDomains parent domains (bar color). Each bar stacks the curated benchmark subset (\benchVideos videos, dark bottom), the YouTube non-benchmark remainder (middle), and the Bilibili (top hatched).}
\vspace{-4mm}
\label{fig:dataset_categories}
\end{figure}

%% file: sections/benchmark.tex
\section{The \KairosBenchEnd}
The dense, time-resolved annotations of \Kairos provide structured evidence for downstream video-language tasks.
We use them to construct \KairosBenchEnd, a benchmark for fine-grained video understanding, by guiding LLM-based question generation with a taxonomy of capabilities and temporal tiers.
Each item is derived from the relevant evidence in the annotation stream and produced as a complete multiple-choice instance, including the question, answer, distractors, and rationale.
The same mechanism can also be adapted to generate instruction-response pairs for downstream training.
Because benchmark construction requires a higher quality standard than raw generation, we further apply a strict verification process before including questions in \KairosBenchEnd.

\vspace{-1mm}
\subsection{Benchmark Design}
\vspace{-1mm}
We design the \KairosBench around three orthogonal axes that are often entangled in long-form video understanding: \emph{Space}, \emph{Time}, and \emph{Dynamics}.
The \emph{Space} axis measures what is present at a single moment, including scenes, entities, spatial relations, on-screen text, and audio cues.
The \emph{Time} axis measures where the required evidence resides and how long the evidence span is, ranging from moment-level cues to whole-video context.
The \emph{Dynamics} axis measures how the video evolves, including within-shot changes, cross-shot continuity, temporal ordering, causal relations, counterfactual reasoning, counting, and holistic integration.
To systematically probe these dimensions, we partition questions into \benchCapabilities capabilities across four cognitive levels: second-level perception, intra-shot evolution, cross-shot reasoning, and whole-video understanding.
Each question is therefore associated with both a capability label and a temporal tier, allowing model performance to be analyzed at a finer granularity.

Given a video, the generator samples evidence only from the timestamped \Kairos annotation stream, and asks \texttt{Gemini-2.5-Flash}~\citep{comanici2025gemini} to produce a complete multiple-choice question (MCQ) in a single call.
This design makes question construction scalable while keeping every question grounded in the same fine-grained evidence used by the dataset.

We design ten source-type samplers, each targeting a different temporal scale and semantic structure in the annotation stream:
\begin{itemize}[leftmargin=*, labelsep=0.4em, itemsep=0pt, topsep=0pt, partopsep=0pt, parsep=0pt]
    \item \textbf{Reference-frame samplers.}
    The \texttt{ref} samplers target local evidence from reference frames.
    \texttt{ref\_perception} samples scene, entity, and spatial information, \texttt{ref\_ocr} samples on-screen text, and \texttt{ref\_audio} samples speech or environmental sound.
    
    \item \textbf{Within-shot samplers.}
    The \texttt{diff} samplers target short-range dynamics within a shot.
    \texttt{diff\_change} samples a reference frame with several subsequent differential descriptions, while \texttt{diff\_sequence} samples the full differential chain of a shot.
    
    \item \textbf{Cross-shot samplers.}
    \texttt{entity\_tracking}, \texttt{transition}, \texttt{cross\_shot}, and \texttt{long\_range} sample entities, events, and transitions across multiple shots, from adjacent-shot changes to multi-minute evidence windows.
    
    \item \textbf{Full-video sampler.}
    \texttt{full\_video} samples evidence over an entire video or a long segment, supporting holistic questions that require extended temporal integration.
\end{itemize}

Each sampler returns the materials needed for question generation, including the relevant annotation descriptions, their timestamps, the evidence span, neighboring reference-frame context, and a pool of candidate distractors.
The evidence span determines the temporal tier of the question: \texttt{T1} for moment-level evidence, \texttt{T2} for evidence within 60 seconds, \texttt{T3} for evidence within 300 seconds, \texttt{T4} for evidence within 900 seconds, and \texttt{T5} for evidence beyond 900 seconds.
At the same time, the source type restricts the possible capability labels from the Space and Dynamics taxonomy.
Thus, the temporal tier and capability label are not assigned post hoc after question generation; they are determined by the evidence sampling process itself.

In the second step, we provide \texttt{Gemini-2.5-Flash}~\citep{comanici2025gemini} with the sampled evidence, the neighboring reference-frame context before and after the target evidence, and a coarse temporal hint indicating the approximate position of the evidence in the video.
The model is required to return a structured response containing four fields: \texttt{question}, \texttt{answer}, \texttt{distractors}, and \texttt{reasoning}.
The question, correct answer, three distractors, and rationale are generated atomically in the same call.
This encourages internal consistency between the answer and the reasoning, and avoids a separate rewriting stage that could make the distractors superficially different from the correct answer.

\noindent\textbf{The distractors design.}
A key design choice is to construct distractors from real annotations rather than hallucinating from scratch.
Specifically, the distractor pool is drawn from non-overlapping time windows of the same video.
Thus, wrong options remain linguistically and semantically plausible because they describe content that actually appears in the video, but they refer to the wrong time point.
This reduces text leakage and prevents models from relying on commonsense priors or eliminating obviously implausible choices.
To answer correctly, a model must locate the relevant temporal evidence and understand the corresponding visual, auditory, or dynamic content.

\vspace{-1mm}
\subsection{Checks of the Benchmark}
\vspace{-1mm}

\noindent\textbf{Leakage control and sanity checks.}
%
%
A common failure mode in video benchmarks is text leakage, where a model can infer the correct answer from the question wording or option priors alone, without actually understanding the video.
We address this issue at both the generation and filtering stages.
At the generation stage, distractors are designed to be plausible rather than fabricated.
%
%
%
This prevents models from answering by simply eliminating obviously implausible choices, instead forces them to locate and understand the relevant visual, auditory, and dynamic evidence. With this design, some questions may still be solvable from textual priors.
We therefore apply a strict text-only audit using a committee of five strong language models: \texttt{GPT-5.4}~\citep{openai2026gpt54}, \texttt{Claude-Opus-4.7}~\citep{anthropic2026claudeopus47}, \texttt{Gemini-3.1-Pro}~\citep{geminiteam2026gemini31pro}, \texttt{GPT-4o-mini}~\citep{openai2026gpt4omini}, and \texttt{Gemma-4-31B-it}~\citep{googledeepmind2026gemma4}.
Each model receives only the question and the four shuffled answer options, without access to the video frames, audio, or textual video annotations.
If at least three of the five models answer a question correctly, the question is marked as text-solvable and removed.
Since each model is evaluated with a single shuffled option order, the random majority floor is approximately 10.4\%.
Across \benchRawMCQs raw questions, this text-only audit removes 19,430 items and leaves \benchAuditedMCQs video-dependent candidates.

\input{fig_table/text_leakage_table}

\noindent\textbf{Human review.}
After the text-only audit, we further introduce a human verification stage to construct the final curated evaluation set.
Each surviving MCQ is independently reviewed by at least two human annotators, who watch the corresponding video and evaluate the item according to four criteria.
First, \textbf{question validity} checks whether the question is clear, unambiguous, and answerable given the video evidence.
Second, \textbf{accuracy and grounding} verifies that the correct answer matches the visual or auditory facts and that the associated temporal evidence is accurate.
Third, \textbf{answer uniqueness} ensures that none of the distractors can reasonably be interpreted as another correct answer.
Fourth, \textbf{video dependence} confirms that the question cannot be answered easily without watching the video. Any item that fails the review criteria is discarded.
For the final leaderboard set, we retain only questions with unanimous pass verdicts, yielding \benchReviewedMCQs human-curated MCQs over \benchVideos videos.
As shown in \tablename~\ref{tab:cross-benchmark-leakage}, we keep the text-only accuracy of \KairosBench substantially lower than that of common video benchmarks, indicating that the retained questions require video-grounded evidence rather than language-only reasoning.

\input{fig_table/benchmark_statistics_image}

\input{fig_table/benchmark_vis}

\vspace{-1mm}
\subsection{Benchmark Statistics}
\vspace{-1mm}
Following the generation, leakage-control, and human-review steps above, the released \KairosBench contains \benchReviewedMCQs human-verified MCQs (and matched Open-ended QA pairs) over \benchVideos videos drawn from all \videoCategories categories.
\figurename~\ref{fig:benchmark_statistics} summarizes the three labeling axes: the \textit{Time} axis is deliberately weighted toward the long-context tail (T3--T5 jointly account for $\approx$30\% of items), the \textit{Source} axis is dominated by reference-frame and within-shot samplers but retains meaningful mass on cross-shot and full-video samplers, and the \textit{Capability} axis spreads over all \benchCapabilities rows so that per-axis evaluation surfaces specific failure modes rather than a single overall score. Representative questions across these axes are shown in \figurename~\ref{fig:benchmark_vis}.

\vspace{-1mm}
\subsection{Results of the \KairosBenchEnd}
\vspace{-1mm}

\input{fig_table/main_results_table}

We evaluate \evalClosedModels closed-source models and \evalOpenModels open-source models. Three models receive the video natively: the two \texttt{Gemini} models and \texttt{LLaVA-Video-7B}. Every other model receives uniformly sampled frames at the per-model budget listed in \tablename~\ref{tab:eval-config}. The results are shown in \tablename~\ref{tab:main_results}. Accuracy drops once the evidence spans several shots or the whole video.

To further analyze text leakage, we evaluate public video-MCQ benchmarks under an identical protocol: we draw a stratified sample of 500 questions from each benchmark and answer them with \texttt{Gemini-3.1-Pro}~\citep{geminiteam2026gemini31pro} using text only, without any visual input. We define leakage as the solver accuracy minus the random-answering baseline for each benchmark. \KairosBench exhibits the lowest leakage despite having the second-highest word count. This is notable because \KairosBench intentionally avoids exposing numeric timestamps that would allow a model to directly retrieve or attend to the referenced segment. Instead, each question contains a linguistic anchor that pins the query to a specific moment in the video while still requiring the model to localize the relevant event from visual content. This design prevents models from taking a timestamp-based shortcut, yet it also makes text-only leakage a more serious concern because the questions must include richer natural-language grounding. To mitigate this risk, \KairosBench undergoes a strict auditing process involving five models followed by human verification.

\input{fig_table/openqa_training_tables}

We additionally run an OpenQA pass on open-source models on \KairosBenchEnd: the model must generate the answer rather than pick a letter. We score with a text-only LLM judge \texttt{Gemini-2.5-Flash}, which gets only question, reference answer, and candidate. It returns an integer 0–3 (no less than 2 counts as correct). The results are shown in \tablename~\ref{tab:openqa_leaderboard}.

\vspace{-1mm}
\subsection{Empowering Video-Language Models with \Kairos}
\vspace{-1mm}

We fine-tune \texttt{Qwen2.5-VL-7B-Instruct} with LoRA for \trainEpochs epoch, on \trainQuestions SFT data (including the question, answer and reasoning) derived from the \Kairos training split. The model is trained using uniformly sampled \trainFrames frames per video. We evaluate on \KairosBench and three public long-video benchmarks (LongVideoBench, LVBench, and Video-MME). Without external training data, the fine-tuned model improves over the base across all benchmarks. The results are shown in \tablename~\ref{tab:training_qwen}, proving that \Kairos serves as a valid supervision target.

%% file: fig_table/text_leakage_table.tex
\begin{table}[t]
\centering
\small
\caption{Using a text-only \texttt{Gemini-3.1-Pro} solver with identical prompts and option shuffling, \KairosBench shows the lowest leakage among all benchmarks, measured by absolute lift over the random baseline, despite having the second-longest question stems.}
\centering
\resizebox{\textwidth}{!}{
\begin{tabular}{lcrrrr}
    \toprule
    Benchmark & \# Options & Random & Accuracy & Leakage $\downarrow$ & \# Avg words $\uparrow$ \\
    \midrule
    CG-Bench~\citep{chen2024cgbenchcluegroundedquestionanswering}       & var 2--8, $\bar N{=}6.86$ & 14.57\% & 50.20\% & +35.63 pp & 48 \\
    DeVE-QA~\citep{Qin_2025}        & fixed 5                   & 20.00\% & 53.60\% & +33.60 pp & 37 \\
    EgoSchema~\citep{mangalam2023egoschema}      & fixed 5                   & 20.00\% & 53.60\% & +33.60 pp & \textbf{154} \\
    LongVideoBench~\citep{wu2024longvideobench} & fixed 4                   & 25.00\% & 56.00\% & +31.00 pp & 87 \\
    Video-Holmes~\citep{cheng2025videoholmesmllmthinklike}   & fixed 6                   & 16.67\% & 47.60\% & +30.93 pp & 54 \\
    Video-MME~\citep{fu2025video}      & fixed 4                   & 25.00\% & 54.80\% & +29.80 pp & 40 \\
    HourVideo~\citep{chandrasegaran2024hourvideo1hourvideolanguageunderstanding}      & fixed 5                   & 20.00\% & 39.00\% & +19.00 pp & 97 \\
    MVBench~\citep{li2024mvbench}        & var 2--5, $\bar N{=}3.51$ & 28.45\% & 41.00\% & +12.55 pp & 32 \\
    LVBench~\citep{wang2025lvbench}        & fixed 4                   & 25.00\% & 37.40\% & \underline{+12.40 pp} & 39 \\
    \midrule
    \Kairos (Ours) & fixed 4 & 25.00\% & 34.60\% & \textbf{+9.60 pp} & \underline{144} \\
    \bottomrule
\end{tabular}}
\label{tab:cross-benchmark-leakage}
\end{table}

%% file: fig_table/benchmark_statistics_image.tex
\begin{figure}[t]
\centering
\includegraphics[width=0.95\textwidth]{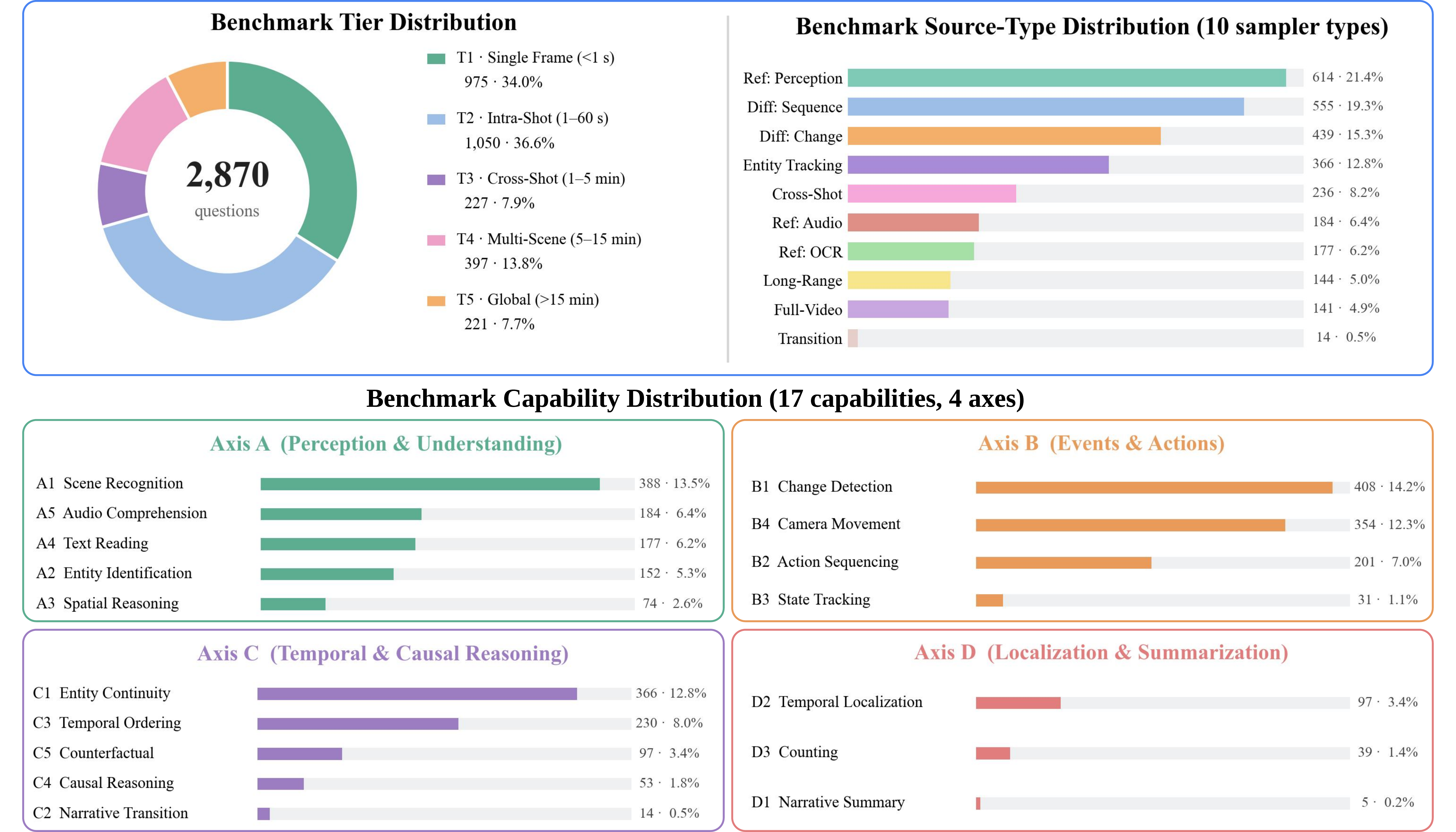}
\caption{Statistics of \KairosBenchEnd. Distribution of the curated \benchReviewedMCQs questions across three independent labeling axes: evidence span, source type, and evaluated capability, showing the benchmark coverage over space, time, and dynamics.}
\label{fig:benchmark_statistics}
\vspace{-4mm}
\end{figure}

%% file: fig_table/benchmark_vis.tex
\begin{figure}[t]
\centering
\includegraphics[width=\textwidth]{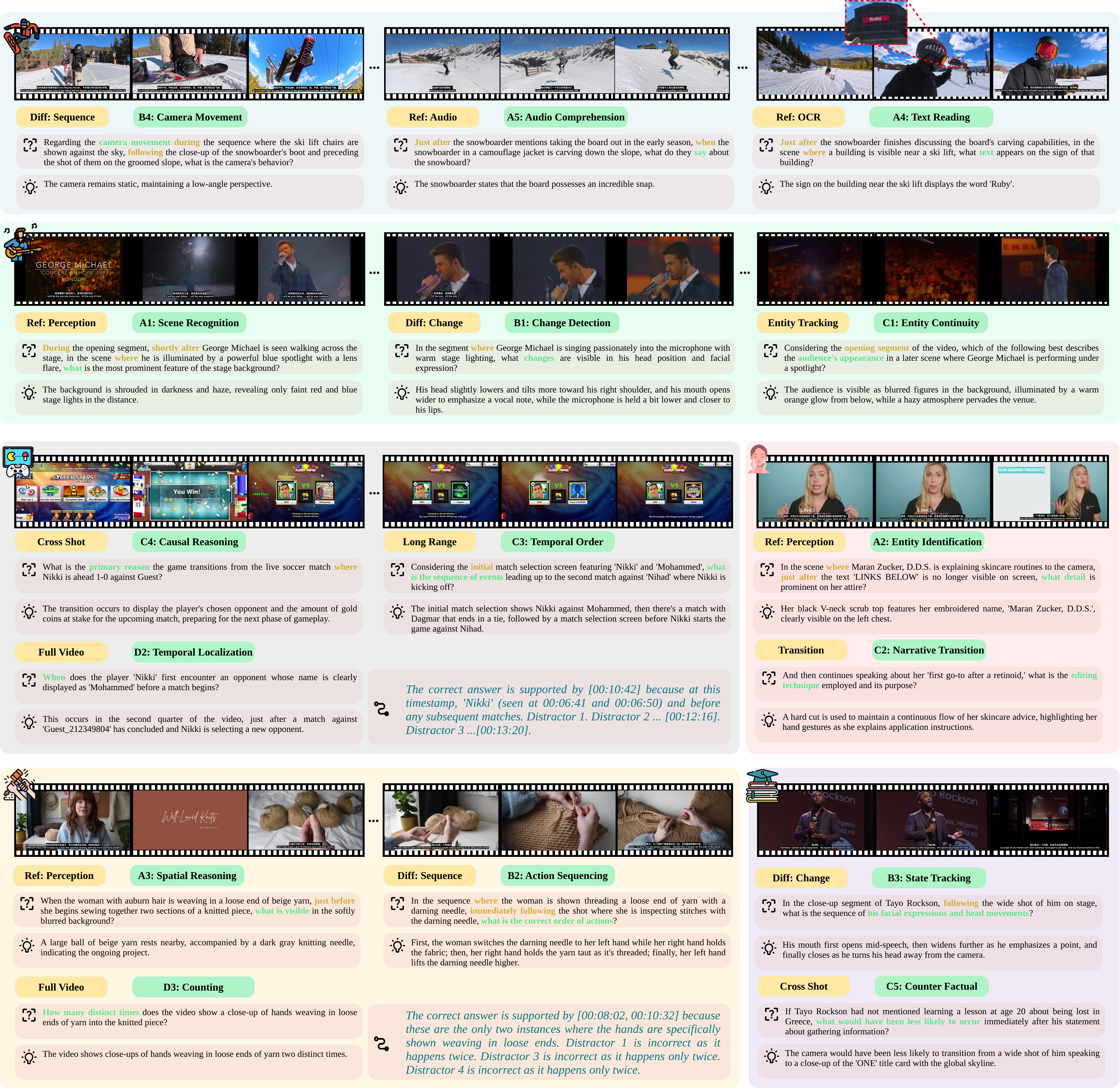}
\vspace{-5mm}
\caption{Representative questions from \KairosBench across the three axes. Each block shows three keyframes, the question and correct answer, and the labels assigned to the question. The examples cover the full evidence span: from single-frame perception, through within-shot evolution and cross-shot continuity, up to full-video.}
\label{fig:benchmark_vis}
\vspace{-6mm}
\end{figure}

%% file: fig_table/main_results_table.tex
\begin{table}[t]
\centering
\caption{MCQ leaderboard on the \KairosBench with full per-capability accuracy. The \benchCapabilities capabilities are grouped by their temporal scope. \textit{\#Q} is the question count per capability.}
\resizebox{\textwidth}{!}{
    \begin{tabular}{l c c ccccc cccc ccccc ccc}
    \toprule
    \multirow{3}{*}{Model} & & \multirow{3}{*}{Overall} & \multicolumn{5}{c}{Single Frame} & \multicolumn{4}{c}{Within Shot} & \multicolumn{5}{c}{Cross Shots} & \multicolumn{3}{c}{Full Video} \\
    \cmidrule(lr){4-8}\cmidrule(lr){9-12}\cmidrule(lr){13-17}\cmidrule(lr){18-20}
     & & & A1 & A2 & A3 & A4 & A5 & B1 & B2 & B3 & B4 & C1 & C2 & C3 & C4 & C5 & D1 & D2 & D3 \\
     & \textit{\#Q} & \footnotesize 2,870 & \footnotesize 388 & \footnotesize 152 & \footnotesize 74 & \footnotesize 177 & \footnotesize 184 & \footnotesize 408 & \footnotesize 201 & \footnotesize 31 & \footnotesize 354 & \footnotesize 366 & \footnotesize 14 & \footnotesize 230 & \footnotesize 53 & \footnotesize 97 & \footnotesize 5 & \footnotesize 97 & \footnotesize 39\\
    \midrule
    \multicolumn{20}{l}{\textcolor{gray}{\textit{Closed-source}}} \\
    [0.5ex]
    \texttt{Gemini-3.1-Pro}~\citep{geminiteam2026gemini31pro} & & \textbf{59.0} & \textbf{61.1} & 54.6 & 67.6 & \textbf{72.3} & \textbf{66.8} & 53.7 & 42.8 & 51.6 & 54.8 & \textbf{64.5} & 14.3 & \textbf{61.3} & \textbf{69.8} & 46.4 & 80.0 & \textbf{81.4} & 30.8 \\
    \texttt{GPT-5.5}~\citep{openai2026gpt55} & & 57.4 & 57.1 & 57.4 & \textbf{78.3} & 64.5 & 39.9 & 55.6 & 46.9 & \textbf{60.9} & \textbf{69.5} & 58.2 & 18.2 & 58.8 & 57.1 & 37.3 & \textbf{100.0} & 74.0 & 35.3 \\
    \texttt{Gemini-2.5-Flash}~\citep{comanici2025gemini} & & 54.1 & 53.4 & \textbf{60.5} & 74.3 & 66.1 & 53.8 & \textbf{56.4} & \textbf{47.3} & 45.2 & 54.2 & 57.1 & 14.3 & 47.0 & 56.6 & 32.0 & 100.0 & 52.6 & 41.0 \\
    \texttt{Nova-2-Lite}~\citep{Intelligence2024} & & 46.8 & 47.7 & 47.4 & 59.5 & 48.6 & 37.0 & 55.9 & 40.3 & 45.2 & 47.2 & 42.9 & \textbf{35.7} & 43.9 & 60.4 & \textbf{48.5} & 100.0 & 39.2 & 33.3 \\
    \texttt{Seed-2.0-Lite}~\citep{bytedanceseed2026seed2} & & 42.6 & 50.8 & 48.0 & 58.1 & 47.5 & 31.5 & 41.7 & 35.8 & 48.4 & 40.4 & 41.8 & 7.1 & 37.4 & 39.6 & 35.0 & 100.0 & 48.5 & \textbf{51.3} \\
    \texttt{GPT-4o}~\citep{achiam2023gpt} & & 37.9 & 42.8 & 43.4 & 50.0 & 44.1 & 31.0 & 32.4 & 32.8 & 25.8 & 39.5 & 42.4 & 21.4 & 33.5 & 34.0 & 28.9 & 100.0 & 42.3 & 28.2 \\
    \texttt{GPT-4o-mini}~\citep{openai2026gpt4omini} & & 31.0 & 32.7 & 27.6 & 32.4 & 32.2 & 23.9 & 35.0 & 29.9 & 41.9 & 24.9 & 35.2 & 21.4 & 31.3 & 41.5 & 27.8 & 80.0 & 19.6 & 41.0 \\
    \midrule
    \multicolumn{20}{l}{\textcolor{gray}{\textit{Open-weight}}} \\
    [0.5ex]
    \texttt{InternVL3-78B}~\citep{chen2024internvl} & & \textbf{52.0} & \textbf{47.9} & \textbf{46.7} & \textbf{64.9} & 53.7 & \textbf{37.0} & \textbf{61.3} & \textbf{45.3} & \textbf{58.1} & 62.4 & \textbf{50.3} & \textbf{50.0} & \textbf{46.1} & \textbf{64.2} & 44.3 & \textbf{100.0} & 47.4 & \textbf{51.3} \\
    \texttt{InternVL3.5-38B}~\citep{wang2025internvl3} & & 47.0 & 46.9 & 46.7 & 62.2 & 50.3 & 29.9 & 53.4 & 35.3 & 38.7 & 54.5 & 42.1 & 50.0 & 41.3 & 62.3 & \textbf{52.6} & 100.0 & \textbf{48.5} & 51.3 \\
    \texttt{GLM-4.5V}~\citep{hong2025glm} & & 46.8 & 44.8 & 44.1 & 54.1 & 53.7 & 34.8 & 50.0 & 39.8 & 51.6 & \textbf{65.8} & 41.3 & 35.7 & 37.4 & 45.3 & 43.3 & 100.0 & 47.4 & 30.8 \\
    \texttt{Qwen3-VL-30B-A3B}~\citep{bai2025qwen3} & & 45.4 & 43.3 & 42.8 & 58.1 & \textbf{54.2} & 28.8 & 46.8 & 37.3 & 41.9 & 63.8 & 42.4 & 28.6 & 38.7 & 54.7 & 41.2 & 100.0 & 41.2 & 30.8 \\
    \texttt{MiMo-VL-7B}~\citep{coreteam2025mimounlockingreasoningpotential} & & 42.3 & 41.5 & 41.4 & 54.1 & 47.5 & 24.5 & 42.2 & 37.8 & 32.3 & 55.1 & 41.8 & 35.7 & 40.4 & 54.7 & 37.1 & 100.0 & 33.0 & 38.5 \\
    \texttt{Qwen3-VL-8B}~\citep{bai2025qwen3} & & 42.2 & 42.0 & 43.4 & 54.0 & 52.5 & 26.1 & 42.6 & 38.3 & 41.9 & 50.6 & 42.1 & 42.9 & 34.4 & 52.8 & 37.1 & 100.0 & 39.2 & 35.9 \\
    \texttt{InternVL3-8B}~\citep{zhu2025internvl3} & & 41.5 & 46.1 & 36.2 & 55.4 & 45.8 & 30.4 & 51.7 & 38.8 & 38.7 & 35.3 & 37.4 & 28.6 & 39.1 & 54.7 & 40.2 & 100.0 & 36.1 & 38.5 \\
    \texttt{Qwen2.5-VL-7B}~\citep{Qwen-VL} & & 41.0 & 37.6 & 40.8 & 54.0 & 44.1 & 35.3 & 42.4 & 39.3 & 41.9 & 48.6 & 38.0 & 35.7 & 37.8 & 45.3 & 43.3 & 100.0 & 36.1 & 33.3 \\
    \texttt{InternVL3.5-8B}~\citep{wang2025internvl3} & & 40.8 & 42.0 & 36.8 & 43.2 & 47.5 & 29.3 & 44.6 & 37.8 & 38.7 & 44.1 & 41.3 & 50.0 & 39.6 & 52.8 & 25.8 & 80.0 & 33.0 & 43.6 \\
    \texttt{Step3-VL-10B}~\citep{step3system} & & 39.7 & 41.5 & 40.1 & 50.0 & 53.7 & 23.9 & 38.7 & 29.9 & 35.5 & 45.2 & 41.5 & 35.7 & 31.7 & 49.1 & 38.1 & 100.0 & 39.2 & 43.6 \\
    \texttt{GLM-4V-9B}~\citep{glm2024chatglm} & & 38.6 & 39.7 & 36.2 & 37.8 & 35.0 & 27.7 & 41.2 & 33.3 & 54.8 & 46.6 & 39.9 & 42.9 & 36.5 & 47.2 & 40.2 & 100.0 & 25.8 & 28.2 \\
    \texttt{Gemma-4-31B}~\citep{googledeepmind2026gemma4} & & 38.2 & 42.3 & 46.7 & 58.1 & 42.9 & 26.1 & 33.1 & 26.9 & 38.7 & 42.1 & 39.3 & 35.7 & 34.8 & 41.5 & 38.1 & 100.0 & 41.2 & 30.8 \\
    \texttt{CogVLM2-Video-13B}~\citep{hong2024cogvlm2} & & 36.3 & 33.8 & 36.8 & 36.5 & 31.1 & 29.4 & 40.2 & 33.3 & 51.6 & 49.4 & 36.9 & 7.1 & 30.9 & 39.6 & 32.0 & 100.0 & 22.7 & 28.2 \\
    \texttt{LLaVA-Video-7B}~\citep{li2024llava} & & 26.8 & 33.0 & 16.4 & 33.8 & 24.3 & 28.8 & 22.1 & 22.4 & 25.8 & 28.0 & 31.1 & 21.4 & 27.8 & 32.1 & 17.5 & 80.0 & 21.6 & 33.3 \\
    \bottomrule
    \end{tabular}%
}
\label{tab:main_results}
\end{table}

%% file: fig_table/openqa_training_tables.tex
\begin{table*}[t]
\centering
\small
\begin{minipage}[t]{0.48\textwidth}
\centering
\vspace{0pt}
\caption{OpenQA leaderboard on the \KairosBenchEnd. \texttt{Gemini-2.5-Flash}~\citep{comanici2025gemini} judge scores each answer against the ground truth on a 0-3 scale, and we report per-tier binarized accuracy with scores no less than 2 counted as correct.}
\resizebox{\linewidth}{!}{
\begin{tabular}{lccccccc}
    \toprule
    Model & & Overall & T1 & T2 & T3 & T4 & T5 \\
     & \#Q & 2,870 & 975 & 1,050 & 227 & 397 & 221 \\
    \midrule
    \texttt{GLM-4.5V}~\citep{hong2025glm} & & \textbf{22.8\%} & 27.2\% & 18.8\% & 20.3\% & \textbf{23.7\%} & 23.5\% \\
    \texttt{MiMo-VL-7B}~\citep{coreteam2025mimounlockingreasoningpotential} & & 22.6\% & \textbf{27.3\%} & \textbf{19.7\%} & \textbf{22.0\%} & 19.1\% & 22.2\% \\
    \texttt{Qwen3-VL-30B-A3B}~\citep{bai2025qwen3} & & 22.4\% & 27.2\% & 19.0\% & 16.3\% & 21.2\% & \textbf{26.7\%} \\
    \texttt{InternVL3-78B}~\citep{chen2024internvl} & & 19.9\% & 26.4\% & 14.1\%          & 17.6\% & 20.2\% & 20.8\% \\
    \texttt{Step3-VL-10B}~\citep{step3system} & & 19.7\% & 25.9\% & 15.7\% & 14.2\% & 18.6\% & 19.0\% \\
    \texttt{Qwen2.5-VL-7B}~\citep{Qwen-VL} & & 19.2\% & 22.8\% & 15.5\% & 15.4\% & 18.9\% & 24.9\% \\
    \texttt{Qwen3-VL-8B}~\citep{bai2025qwen3} & & 19.2\% & 25.9\% & 12.4\% & 14.1\% & 21.7\% & 22.2\% \\
    \texttt{InternVL3.5-38B}~\citep{wang2025internvl3} & & 16.9\% & 22.8\% & 11.5\% & 15.9\% & 18.6\% & 14.0\% \\
    \texttt{InternVL3.5-8B}~\citep{wang2025internvl3} & & 14.8\% & 20.8\% & 9.6\% & 13.2\% & 15.1\% & 14.5\% \\
    \texttt{InternVL3-8B}~\citep{zhu2025internvl3} & & 14.7\% & 21.0\% & 9.4\% & 13.2\% & 14.1\% & 14.9\% \\
    \texttt{GLM-4V-9B}~\citep{glm2024chatglm} & & 10.8\% & 16.3\% & 6.6\% & 8.4\% & 8.6\% & 13.1\% \\
    \texttt{LLaVA-Video-7B}~\citep{li2024llava} & & 8.3\% & 9.6\% & 3.2\% & 6.2\% & 14.4\% & 17.6\% \\    
    \bottomrule
    \end{tabular}
}
\label{tab:openqa_leaderboard}
\end{minipage}
\hfill
\begin{minipage}[t]{0.48\textwidth}
\centering
\vspace{0pt}
\caption{
\Kairos training data improves \texttt{Qwen2.5-VL-7B-Instruct}~\citep{Qwen-VL} on \KairosBench and three external long-video benchmarks via LoRA~\citep{hu2022lora} fine-tuning. At evaluation stage, we ablate the frame budget over $\{16,32,64\}$. Without external training data, the fine-tuned model improves over the base across all benchmarks.}

\resizebox{\linewidth}{!}{
\begin{tabular}{lcccc}
    \toprule
    \multirow{2}{*}{Model} & \multicolumn{4}{c}{Benchmark} \\
    \cmidrule(lr){2-5}
     & \Kairos & LongVideoBench & LVBench & Video-MME \\
     \texttt{base} (32-f) & 40.42 & 56.29 & 38.69 & 57.11 \\
     \texttt{finetuned} (16-f) & \textbf{47.94} & 58.00 & 39.18 & 57.63 \\
     \texttt{finetuned} (32-f) & 47.49 & \textbf{60.08} & 40.81 & \textbf{59.56} \\
     \texttt{finetuned} (64-f) & 45.12 & 59.50 & \textbf{42.15} & 59.30 \\
    \bottomrule
\end{tabular}
}
\label{tab:training_qwen}
\end{minipage}
\vspace{-2mm}
\end{table*}

%% file: sections/conclusion.tex
\section{Conclusion}

We present \KairosEnd, a dataset of \dataVideos long-form videos with 1 FPS temporally grounded annotations, and \KairosBenchEnd, a strictly audited benchmark of \benchReviewedMCQs MCQ and OpenQA questions organized along three orthogonal axes (Space, Time, Dynamics).
\KairosBench is the cleanest of nine public video-MCQ benchmarks under an identical text-only-leakage probe.
Fine-tuning a 7B open-weight VLM on \trainQuestions SFT data corpus derived from \Kairos dataset improves accuracy on \Kairos and on three external long-video benchmarks despite using no training data from them.
We release the dataset, benchmark, and pipeline to enable evaluation and supervision of long-form video understanding at the granularity at which it actually unfolds.

%% file: sections/relatedworks.tex
\section{Related Work}
\paragraph{Video Understanding and Video-Language Models.}
Early video understanding relied on hand-crafted spatiotemporal descriptors and trajectory-based representations \citep{laptev2005space,klaser2008spatio,wang2013action,wang2013dense}.
These methods established the fundamentally spatiotemporal nature of video analysis, but they typically produced clip-level decisions rather than persistent representations of entities and states over time.
Deep architectures later shifted the field toward learnable spatiotemporal features \citep{tran2015learning,carreira2017quo,feichtenhofer2019slowfast,wang2016temporal,zhou2018temporal,wang2018non}.
While these models substantially improved video representations, their supervision was still largely aligned with clips or full videos rather than explicitly with second-level state transitions or temporally consistent entity attributes. 
Video Transformers further improved long-range temporal modeling through native cross-frame attention \citep{bertasius2021space,arnab2021vivit,liu2022video}, while self-supervised pretraining methods demonstrated strong transfer without dense annotation \citep{tong2022videomae,wei2022masked,akbari2021vatt,wang2022internvideo}.
Nevertheless, much of this literature still optimizes coarse temporal objectives and often relies on sparse temporal sampling or compressed visual tokens for efficiency, which can weaken sensitivity to subtle state changes in long videos.

Video-language research evolved from sequence-to-sequence captioning models such as S2VT \citep{venugopalan2015sequence} to large-scale video-text pretraining frameworks \citep{sun2019videobert,zhu2020actbert,li2020hero,luo2020univl,fu2021violet,li2022align,xu2021videoclip}.
Large web-scale corpora such as HowTo100M \citep{miech2019howto100m} and WebVid \citep{bain2021frozen}, together with CLIP-style contrastive learning \citep{radford2021learning}, enabled more scalable cross-modal alignment and inspired retrieval-oriented extensions \citep{luo2021clip4clip,luo2022clip4clip,ma2022x,fang2021clip2video,bain2021frozen}.
However, these corpora are often only weakly aligned in time, and most training objectives emphasize global or clip-level matching rather than second-level, temporally anchored scene grounding. 
Recent Video-LLMs connect visual encoders to large language models through adapters, query modules, or projection layers.
Connector-based architectures such as Flamingo \citep{alayrac2022flamingo} and BLIP-2 \citep{li2023blip} established scalable multimodal interfaces, and subsequent systems, including Video-LLaMA \citep{zhang2023video}, Video-ChatGPT \citep{maaz2024video}, LLaVA-Video~\citep{li2024llava}, LLaMA-VID \citep{li2024llama}, and TimeChat \citep{ren2024timechat}, extended this paradigm to video dialogue, instruction following, and long-context reasoning.
Even so, a central tension remains between temporal coverage and spatiotemporal fidelity: sparse frame sampling, frame pooling, and aggressive token reduction improve throughput, but they can hinder fine-grained temporal localization, entity tracking, and temporally consistent reasoning.

\paragraph{Video-Language Datasets and Benchmarks.}
Large-scale video-text pretraining datasets provide the foundation for many modern video-language models.
Web-scale corpora such as WebVid \citep{bain2021frozen}, InternVid \citep{wang2023internvid}, and OpenVid-1M \citep{nan2024openvid} collect millions to hundreds of millions of video-text pairs, enabling scalable video-text representation learning.
These datasets improve coverage and diversity, but their annotations are weakly aligned and lack dense temporal supervision.
More recent high-quality captioning datasets, such as ShareGPT4Video \citep{chen2024sharegpt4video}, and FineVideo~\citep{Farre2024FineVideo}, move toward richer and more structured supervision.
Another important line of work focuses on video instruction tuning and conversational video data.
Datasets such as VideoChat-11K \citep{li2025videochat}, Video-ChatGPT-100K \citep{maaz2024video}, Valley-Instruct-65K \citep{luo2023valley}, LLaVA-Video-178K \citep{zhang2024llava}, ShareGPTVideo \citep{chen2024sharegpt4video}, and TimeIT \citep{ren2024timechat} transform existing captions, QA annotations, or video understanding tasks into instruction-following formats.
While these datasets provide richer language supervision, much of their annotations are still coarse in spatial and temporal granularity. As a result, they lack fine-grained grounding in space and time and provide limited supervision over video dynamics.

Temporal grounding and dense video understanding benchmarks provide more explicit temporal supervision.
Classic and widely used datasets such as ActivityNet \citep{krishna2017dense}, Charades-STA \citep{gao2017tall}, DiDeMo \citep{anne2017localizing}, and QVHighlights \citep{lei2021detecting} annotate the relationship between natural language queries and temporal segments in videos.
These benchmarks have played an important role in moving video-language evaluation beyond full-video classification and toward timestamp-aware grounding.
More recent long-form or egocentric grounding resources, such as MAD \citep{soldan2022mad} and Ego4D \citep{grauman2022ego4d}, extend this setting to movies or first-person videos.
VidChapters-7M \citep{yang2023vidchapters} further introduces large-scale chapter-level supervision for long videos, with chapter titles and timestamps collected from web videos.
These datasets are highly relevant to temporal localization, but they still do not fully solve the problem of fine-grained state and attribute tracking.
Moment boundaries are often coarse, query-level, or segment-level, and the annotations usually describe events or steps rather than maintain a persistent inventory of entities, attributes, relations, and state changes across time.

Video question-answering benchmarks have also evolved from short-clip QA toward long-form and diagnostic evaluation.
Earlier benchmarks evaluate video understanding through manually annotated or carefully constructed questions, including both open-ended and multiple-choice formats~\citep{yu2019activitynet,xiao2021next,grauman2022ego4d}.
More recent comprehensive benchmarks, including MVBench \citep{li2024mvbench}, LongVideoBench \citep{wu2024longvideobench}, Video-MME \citep{fu2025video}, and LVBench \citep{wang2025lvbench}, expand evaluation toward multi-task reasoning, long-video understanding, and full-spectrum video modeling.
These benchmarks have been crucial for revealing the limitations of Video LLMs under long-context settings, but most of them still reduce evaluation to discrete multiple-choice accuracy.
This makes leaderboards easy to compare, but it can also obscure whether a model truly tracks temporal evidence, maintains entity consistency, or merely exploits language priors and coarse scene summaries.

Several diagnostic and multi-task benchmarks aim to evaluate more detailed perceptual and temporal capabilities.
The Perception Test includes object tracks, point tracks, action segments, sound segments, multiple-choice video QA, and grounded video QA, providing a more diverse testbed for perception, grounding, and multimodal reasoning~\citep{patraucean2023perception}.
Datasets such as STAR, CLEVRER, EgoTaskQA, and TGIF-QA focus on situated reasoning, physical reasoning, egocentric reasoning, state transitions, and repetition counting~\citep{wu2021star_situated_reasoning,CLEVRER2020ICLR,jia2022egotaskqa,jang2019video}.
These datasets isolate specific reasoning factors but are typically built on short or synthetic clips and thus complement rather than replace long-video benchmarks that require sustained temporal reasoning.

Overall, existing datasets and benchmarks have substantially advanced video-language learning, but they leave an important gap for fine-grained, temporally faithful video understanding. Large-scale web corpora provide breadth but only weak temporal and spatial alignment. Instruction datasets improve the conversational interface of VideoLLMs but often rely on synthetic or model-generated supervision. Temporal grounding datasets provide timestamp supervision but usually focus on query-to-moment localization rather than persistent entity and state modeling. Long-video QA benchmarks expose the difficulty of reasoning over extended contexts, but their multiple-choice format can underdiagnose perceptual failures, temporal hallucinations, and entity-state inconsistencies. These limitations suggest the need for benchmarks and training resources that combine long temporal coverage with dense temporal anchoring, explicit spatial and entity-level grounding, and evaluation protocols that test not only whether a model gives the correct answer, but also whether it can maintain temporally consistent, evidence-grounded representations throughout the video.


%% file: sections/appendix.tex
\newcommand{\dataYouTubeHours}{2,857~}
\newcommand{\dataDurationMedianMin}{17.0~}
\newcommand{\dataDurationMeanMin}{18.2~}
\newcommand{\dataDurationMaxMin}{32.7~}
\newcommand{\dataShotsTotal}{1,192,552~}
\newcommand{\dataShotsPerVideoMedian}{91~}
\newcommand{\dataShotsPerVideoMean}{126.4~}
\newcommand{\dataShotsPerVideoMax}{1,235~}
\newcommand{\dataShotDurMedianSec}{3.7~}
\newcommand{\dataShotDurMeanSec}{8.6~}
\newcommand{\dataAnnotateBatches}{16~}
\newcommand{\dataEntityTypes}{8~}

\newcommand{\pipeFps}{1~}
\newcommand{\pipeShotThreshold}{0.5~}
\newcommand{\pipeShotMinLen}{15~}
\newcommand{\pipeMaxImageDim}{1024~}
\newcommand{\pipeJpegQuality}{90~}
\newcommand{\pipeReanchorInterval}{300~}
\newcommand{\pipeNarrativeWindow}{20~}
\newcommand{\pipeDiffWindow}{50~}
\newcommand{\pipeTemperature}{0.2~}
\newcommand{\pipeTensorParallel}{2~}
\newcommand{\pipeGpuMemUtil}{0.85~}
\newcommand{\pipeBatchTokens}{1.2M~}
\newcommand{\pipeRestartRounds}{150~}
\newcommand{\pipeAudioWindowSec}{30~}
\newcommand{\pipeRosterCap}{400~}

\newcommand{\srcRefPerception}{614~}
\newcommand{\srcRefAudio}{184~}
\newcommand{\srcRefOcr}{177~}
\newcommand{\srcDiffChange}{439~}
\newcommand{\srcDiffSequence}{555~}
\newcommand{\srcEntityTracking}{366~}
\newcommand{\srcCrossShot}{236~}
\newcommand{\srcLongRange}{144~}
\newcommand{\srcTransition}{14~}
\newcommand{\srcFullVideo}{141~}


\newcommand{\reviewAnnotators}{34~}
\newcommand{\reviewPassConsensus}{2,876~}
\newcommand{\reviewUnanimousPass}{2,870~}
\newcommand{\reviewLeakageProbe}{500~}

\newcommand{\leakageKairosRawAcc}{36.6\%}
\newcommand{\leakageKairosRawLift}{$+11.6$\,pp~}

\newcommand{\evalDecode}{greedy~}
\newcommand{\evalMaxImagesPerQ}{16~}
\newcommand{\evalNativeVideoModels}{\textsc{gemini-2.5-flash}, \textsc{gemini-3.1-pro}, \textsc{llava-video-7b}~}

\newcommand{\ftBaseModel}{\textsc{Qwen2.5-VL-7B-Instruct}~}
\newcommand{\ftLoraRank}{32~}
\newcommand{\ftLoraAlpha}{64~}
\newcommand{\ftCutoffLen}{16,384~}
\newcommand{\ftLR}{$1\!\times\!10^{-4}$~}
\newcommand{\ftScheduler}{cosine~}
\newcommand{\ftWarmup}{0.03~}
\newcommand{\ftEffectiveBatch}{32~}
\newcommand{\ftPrecisionFT}{bf16~}
\newcommand{\ftHardwareFT}{4$\times$H200~}
\newcommand{\ftFrameworkFT}{LLaMA-Factory~}
\newcommand{\ftFrames}{32~}
\newcommand{\ftEpochsFT}{1~}
\newcommand{\ftTrainShards}{6~}
\newcommand{\ftSourceVideos}{1,870~}

\newcommand{\pipeVLM}{\textsc{Qwen3-VL-8B-Instruct}~}
\newcommand{\pipeASR}{\textsc{faster-whisper-large-v3}~}
\newcommand{\pipeEnvAudio}{\textsc{Qwen2-Audio-7B-Instruct}~}
\newcommand{\pipeShot}{\textsc{TransNetV2}~}
\newcommand{\pipeGen}{\textsc{Gemini-2.5-Flash}~}
\newcommand{\throughputGpuHours}{715~}

\newcommand{\benchTone}{975~}
\newcommand{\benchTtwo}{1,050~}
\newcommand{\benchTthree}{227~}
\newcommand{\benchTfour}{397~}
\newcommand{\benchTfive}{221~}

\newcommand{\auditPanel}{5~}
\newcommand{\auditThreshold}{3~}
\newcommand{\auditRemoved}{19,430~}
\newcommand{\auditRandomFloor}{10.4\%}

\newcommand{\leakageKairos}{$+9.6$\,pp~}
\newcommand{\leakageKairosAcc}{34.6\%}
\newcommand{\leakageSolver}{\textsc{gemini-3.1-pro}~}

\newcommand{\openqaJudge}{\textsc{gemini-2.5-flash}~}
\newcommand{\openqaThreshold}{$\ge\!2$~}

\newcommand{\ftDDP}{DeepSpeed ZeRO-3~}

%

\section{Annotation Pipeline Hyperparameters}
\label{app:pipeline}

This section enumerates the hyperparameters used by every pipeline stage so that the corpus and the benchmark can be reproduced exactly.
The four artifacts emitted per video are described in Table~\ref{tab:per-video-artifacts} of \S\ref{app:dataset}; the present section concerns \emph{how} those artifacts are produced.

\subsection{Shot Detection and Frame Sampling}

\noindent\textbf{Shot detector.}
Shot boundaries are produced by \pipeShot~\citep{soucek2024transnet} with a softmax threshold of \pipeShotThreshold and a minimum scene length of \pipeShotMinLen source frames.
The threshold is intentionally slightly recall-biased: a missed boundary merges two semantically distinct shots and breaks every cross-shot question generated for the resulting merged span, whereas a spurious boundary at most introduces a redundant reference frame inside a real shot.
Both hard cuts and gradual transitions are accepted as boundaries; the gradual-transition probability head is read from the same TransNetV2 forward pass.

\noindent\textbf{Frame sampling.}
After shot detection, the pipeline materialises a per-frame structural record and extracts a \pipeFps FPS JPEG subset for downstream annotation.
Frames are resized so that the longer image dimension is \pipeMaxImageDim pixels, and saved at JPEG quality \pipeJpegQuality.
The resulting \pipeFps FPS timeline is the spine of the annotation stream and is what every subsequent description, audio segment, and benchmark evidence span is keyed to.

\noindent\textbf{Reference and differential frames.}
Within each shot, the first extracted frame is treated as the \emph{initial frame} and receives a full spatial description plus, when entity tracking is enabled, a structured entity list.
All later \pipeFps FPS samples in the same shot are treated as \emph{differential frames} and are described only in terms of what changed relative to the previous state.
For long shots, the pipeline re-anchors every \pipeReanchorInterval differential frames by promoting the current frame to a new initial frame, which keeps prompt context bounded and reduces drift in long description chains.

\subsection{Audio Pipeline}

\noindent\textbf{Two complementary audio models.}
Audio is processed by two complementary models running on the same physical device.
Speech is transcribed with \pipeASR~\citep{faster_whisper, radford2023robust} at sentence resolution.
Non-speech audio is summarized by \pipeEnvAudio~\citep{chu2024qwen2} over fixed \pipeAudioWindowSec-second windows; this captures environmental and event-level sounds such as crowd noise, engine sounds, applause, footsteps, and music.

\noindent\textbf{Alignment.}
The two audio streams are aligned to the same \pipeFps FPS timeline used by the visual annotation stage.
Each speech sentence is attached to the first processed frame whose timestamp falls inside the sentence's time span, and never attached twice.
Environmental audio summaries are attached only to initial frames, not to every differential frame, so that nearby records are not burdened with repeated ambient descriptions.

\subsection{Visual Annotation Prompts}

\noindent\textbf{Server.}
The visual annotation stage is driven by \pipeVLM served through vLLM~\citep{kwon2023efficient} with tensor parallel size \pipeTensorParallel, \texttt{bfloat16} weights, sampling temperature \pipeTemperature, and \texttt{gpu\_memory\_utilization} \pipeGpuMemUtil.

\noindent\textbf{Four prompt templates.}
The pipeline emits descriptions through four prompt templates:
\begin{itemize}[leftmargin=*, labelsep=0.4em, itemsep=0pt, topsep=0pt, partopsep=0pt, parsep=0pt]
    \item \texttt{NARRATIVE\_REF\_WITH\_ENTITIES} (initial frames). Returns structured JSON of the form \texttt{\{"description": ..., "entities": [...]\}}: a dense paragraph plus an entity list with canonical mention, entity type, and a \texttt{visual\_details} string for re-identification. Conditioned on the current frame, the current entity bank, and a narrative context window of the previous \pipeNarrativeWindow shot descriptions.
    \item \texttt{DIFFERENTIAL} (within-shot differential frames). Receives the previous and current frame plus a shot-local description chain capped at the previous \pipeDiffWindow descriptions. Forbids restating already-described static content; allows an empty string when no meaningful change is observed.
    \item \texttt{TRANSITION} (shot boundaries). Receives the two boundary frames of adjacent shots together with their shot descriptions, and asks the model to describe the cut in terms of editing technique, narrative purpose, and framing change without restating scene content already covered.
    \item \texttt{REFERENCE} (fallback). A plain single-frame captioning path used when entity-structured output is not needed.
\end{itemize}

\noindent\textbf{Audio injection.}
Aligned ASR and environmental audio are appended to every prompt as a shared suffix that contains only the not-yet-used segments anchored to the current timestamp; this brings spoken and ambient content into the same frame-level annotation stream without a separate fusion stage.

\subsection{Cross-shot Entity Matching}

\noindent\textbf{Entity vocabulary.}
For each initial frame the model emits a structured set of entities drawn from the fixed vocabulary of \dataEntityTypes categories: \texttt{person}, \texttt{animal}, \texttt{object}, \texttt{vehicle}, \texttt{text}, \texttt{location}, \texttt{food}, \texttt{clothing}.
Every instance carries a canonical phrase and a concise \texttt{visual\_details} string intended to preserve identifying appearance cues across shots.

\noindent\textbf{Label-exact matching.} Cross-shot entity assignment relies on the prompt contract rather than on a similarity threshold. The current bank is rendered into each initial-frame prompt as one record per entity (canonical name, type, and \texttt{visual\_details}), limited to the \pipeRosterCap most recently seen entities, and the model is instructed to reuse a bank entity's canonical name verbatim. An emitted mention is matched against the bank by normalized label equality: lower-casing, removal of a leading article and of edge punctuation, and whitespace collapsing. A hit appends an appearance (shot, frame, surface phrase) to the existing entity; a miss registers a new entity ID.

\subsection{Mega-batch Inference}

\noindent\textbf{Single batch across the workload.}
Each inference round collects the initial-frame prompts of at most one shot per video (a long shot re-anchored every \pipeReanchorInterval frames contributes one prompt per segment) and then fills the remaining token budget with differential prompts drawn from segments whose initial frame has already been completed.
Prompt lengths are estimated analytically before batching and any prompt that would overflow the round budget is deferred.
The resulting workload is submitted through one \texttt{vllm.generate\_batch} call.

\noindent\textbf{Token budget and engine cadence.}
The hard cap is \pipeBatchTokens prompt tokens per round, large enough to saturate both inference GPUs while remaining stable against out-of-memory spikes from heavy multi-image prompts.
The vLLM engine is rebuilt every \pipeRestartRounds rounds so that shared-memory artifacts that accumulate over long tensor-parallel sessions do not destabilise multi-hour annotation jobs.

\subsection{Hardware and Throughput}

The pipeline runs on a homogeneous H200 cluster.
The \texttt{prep} and \texttt{audio} stages each exceed $20\times$ real time on a single H200, while the \texttt{infer} stage reaches roughly $4\times$ real time per H200, or $8\times$ aggregate throughput under tensor-parallel-2 serving.

\section{Per-video Schema and Distribution Diagnostics}
\label{app:dataset}

\subsection{Per-video Schema}

Every video directory holds the six artifacts listed in Table~\ref{tab:per-video-artifacts}.
The central supervision channel is \texttt{descriptions.jsonl}: one JSON record per \pipeFps FPS extracted frame, containing the absolute timestamp, the shot identifier and shot start/end frames and times, a flag \texttt{is\_reference} distinguishing reference frames from differential frames, the description text, and, on reference frames that sit at a shot boundary, a populated \texttt{transition} field carrying the cross-shot dynamics description.
This single record type is sufficient to support every axis of the derived benchmark and every granularity of the fine-tuning corpus --- a downstream sampler need only filter by \texttt{is\_reference} and the presence of \texttt{transition} to obtain the artifact it wants.

\begin{table}[t]
\centering
\small
\caption{\textbf{Per-video output schema.} \texttt{descriptions.jsonl} is the central supervision artifact; the other five files carry structural metadata that supports downstream sampling.}
\label{tab:per-video-artifacts}
\setlength{\tabcolsep}{4pt}
\begin{tabular}{@{}p{0.27\textwidth} l p{0.55\textwidth}@{}}
\toprule
\textbf{File} & \textbf{Producer} & \textbf{Role} \\
\midrule
\texttt{frames/\{frame\_id\}.jpg}       & \texttt{prep}  & \pipeFps FPS JPEG samples, $\le$\pipeMaxImageDim\,px, JPEG quality \pipeJpegQuality. \\
\texttt{prep\_cache.json}               & \texttt{prep}  & shot list and frame manifest, fingerprinted by the shot-detector and sampler config. \\
\texttt{audio\_segments.json}           & \texttt{audio} & \pipeASR speech segments and \pipeEnvAudio environment summaries. \\
\texttt{descriptions.jsonl}             & \texttt{infer} & per-frame reference and differential descriptions; \texttt{transition} field on shot boundaries. \\
\texttt{entities\_final.json}           & \texttt{infer} & per-video entity bank: canonical label, type, visual details, first appearance. \\
\texttt{resume\_state.json}             & \texttt{infer} & entity counter and ASR deduplication state; checkpointed at every shot boundary. \\
\bottomrule
\end{tabular}
\end{table}

\subsection{Source Curation}

\noindent\textbf{Two-platform pool, three-level taxonomy.}
Videos are drawn from two public long-form video platforms, YouTube and Bilibili, seeded by two curated download lists that together pool \videoURLs candidate URLs (\videoYouTubeURLs YouTube + \videoBilibiliURLs Bilibili).
Each row is pre-tagged with a three-level content taxonomy: \videoDomains parent \emph{domains} (\texttt{A. Sports}, \texttt{B. Gaming}, \texttt{C. Ego-Centric \& Daily Life}, \texttt{D. Vlogs \& Ceremonies}, \texttt{E. Arts \& Crafts}, \texttt{F. Media \& Entertainment}, \texttt{G. Public Safety}, \texttt{H. Embodied AI}, \texttt{I. Drones \& Remote Sensing}, \texttt{J. AIGC-related Content}, \texttt{K. Formal Communication}, \texttt{L. Computer Use}); \videoCategories \emph{categories} nested within domains; and \videoScenarios \emph{scenarios} at the leaf level (e.g.\ Basketball within \texttt{A. Sports / I. Ball Games}).
The full taxonomy --- every leaf scenario under its parent (domain, category) --- is enumerated in Table~\ref{tab:full-taxonomy}.

{\small
\setlength{\tabcolsep}{4pt}
\renewcommand{\arraystretch}{1.05}
\begin{longtable}{@{}p{2.4cm} p{2.7cm} p{8.0cm}@{}}
\caption{\textbf{Full \videoDomains-domain / \videoCategories-category / \videoScenarios-scenario content taxonomy of \KairosEnd.} Each candidate URL in the \videoURLs-URL pool was pre-tagged with one (domain, category, scenario) triple at curation time. The same taxonomy carries through to the annotation corpus and to the benchmark sampler, so any per-axis evaluation cut can be re-grouped by content type.}
\label{tab:full-taxonomy}\\
\toprule
\textbf{Domain} & \textbf{Category} & \textbf{Scenarios} \\
\midrule
\endfirsthead
\multicolumn{3}{l}{\emph{(\,...continued from previous page)}}\\
\toprule
\textbf{Domain} & \textbf{Category} & \textbf{Scenarios} \\
\midrule
\endhead
\midrule
\multicolumn{3}{r}{\emph{(continued on next page...)}} \\
\endfoot
\bottomrule
\endlastfoot
\textbf{A. Sports} & I. Ball Games & Basketball, Soccer/Football, Volleyball, Baseball, American Football, Golf, Snooker, Bowling \\
 & II. Racket Sports & Tennis, Badminton, Table Tennis \\
 & III. Water \& Ice Sports & Swimming, Diving, Water Polo, Ice Hockey, Curling, Figure Skating, Speed Skating, Short Track \\
 & IV. Athletics & Racing, Relay Race, Marathon, Long Jump, High Jump, Hurdles, Pole Vault, Shot Put, Discus Throw, Javelin Throw, Hammer Throw \\
 & V. Gymnastics & Artistic Gymnastics, Trampoline \\
 & VI. Combat Sports & Boxing, Wrestling, Judo, Taekwondo, Karate, MMA, WWE, Kickboxing, Wushu, Fencing, Sumo \\
 & VII. Racing Sports \& Equestrian & Formula 1, Rally Racing, Off-road Racing, Motorcycle Racing, Horse Racing, Equestrian \\
 & VIII. Extreme Sports & Skateboarding, BMX, Parkour, Surfing, Alpine Skiing, Snowboarding, Rock Climbing, Bouldering, Skydiving, Bungee Jumping, Scuba Diving \\
\cmidrule(l){1-3}
\textbf{B. Gaming} & I. Video Games & First-Person Shooter, MOBA, VR Games, Soulslike, Roguelike, Sandbox Games, Racing Simulation, Sports Simulation, Horror Games, Puzzle Games \\
 & II. Board \& Strategy Games & Go, Chess, Chinese Chess, Gomoku, Mahjong, Poker, Bridge \\
\cmidrule(l){1-3}
\textbf{C. Ego-Centric \& Daily Life} & I. Household Activities & Cooking, Washing Dishes, Folding Laundry, Ironing Clothes, Vacuuming, Assembling Furniture, Fixing Furniture, Cleaning, Child Care \\
 & II. Daily Skills & Typing, Handwriting, Tool Using, Equipment Operation, Playing Instruments, Conversation, Studying, Teaching, First Aid, Car Repair, Tire Change \\
 & III. Personal Care & Hand Washing, Makeup, Skincare, Taking Medicine, Rehabilitation \\
 & IV. Outdoor Activities & Gardening, Agricultural Labor, Tourism \\
\cmidrule(l){1-3}
\textbf{D. Vlogs \& Ceremonies} & I. Vlogs & Daily Vlogs, Travel Vlogs, Walking Vlogs, Running Vlogs, Riding Vlogs, Shopping Vlogs, Gym Vlogs \\
 & II. Ceremonies & Wedding Ceremony, Birthday Party \\
\cmidrule(l){1-3}
\textbf{E. Arts \& Crafts} & I. Visual Arts & Pencil Sketching, Oil Painting, Watercolor Painting, Calligraphy \\
 & II. Handcraft & Pottery, Origami, Paper Cutting, Embroidery, Knitting, Woodworking, Sculpting, Jewelry Making, Glass Blowing, Leather Crafting, 3D Printing \\
\cmidrule(l){1-3}
\textbf{F. Media \& Entertainment} & I. Drama Genres & Medical, Legal, Crime, Domestic, School, Xianxia, Spy, Office \\
 & II. Shows \& Performance & Talk Show, Stand-up Comedy, Magic, Street Performance, Circus, Acrobatics, Concert \\
 & III. News \& Documentary & News Broadcast, Documentary \\
 & IV. Musical \& Opera & Musical, Opera \\
 & V. Animation & Animated Films, Animated Series \\
\cmidrule(l){1-3}
\textbf{G. Public Safety} & I. Driving & Urban Driving, Highway Driving \\
 & II. Surveillance & Traffic Surveillance, Public Space Monitoring, Home Security \\
\cmidrule(l){1-3}
\textbf{H. Embodied AI} & I. Embodied Interaction & Manipulation, Tool Use, Navigation, Human-Robot Interaction, Dexterous Manipulation \\
\cmidrule(l){1-3}
\textbf{I. Drones \& Remote Sensing} & I. Aerial Video & Aerial Footage, Survey Flights \\
 & II. Satellite Video & Satellite Video, Time-lapse Observation \\
\cmidrule(l){1-3}
\textbf{J. AIGC-related Content} & I. Generated Content & Text-to-Video Samples, AI Stylized Animation, Deepfake \\
 & II. Artifacts \& Consistency & Compression Artifacts, Moir\'e, AI Motion Glitch, Frame Interpolation Artifacts, Temporal Consistency Issues \\
\cmidrule(l){1-3}
\textbf{K. Formal Communication} & I. Academic & Conference Presentation, Plenary Speech, Poster Presentation, Job Talk, Seminar, Group Meeting, Dissertation Defense, TED-style Talk, Lecture \\
 & II. Business & Investor Pitch, Product Launch, Board Meeting, Contract Negotiation \\
 & III. Politics & Stump Speech, Acceptance Speech, State of the Union, Press Conference, Legislative Debate, Diplomatic Negotiation, Parliamentary Session \\
\cmidrule(l){1-3}
\textbf{L. Computer Use} & I. Software Tutorial & Word, Excel, PowerPoint, PhotoShop, LightRoom, Premiere, Blender, VS Code \\
 & II. Daily Activities & Browse, E-mail \\
\end{longtable}
}

\noindent\textbf{Duration filtering.}
Candidate URLs are filtered to the \dataMinMinutes--\dataMaxMinutes minutes duration band: the length regime where clip-scale benchmarks stop scaling and where the Time axis becomes non-trivial.
Sources longer than \dataMaxMinutes min are split into consecutive fixed-length \dataMaxMinutesEnd-minute parts (the \texttt{\_partNN} suffixes visible in the corpus) rather than truncated, so no footage is discarded.
After duration filtering and download-side failures (privacy takedowns, geo-restrictions, channel deletion), the final annotation corpus contains \dataVideos videos totalling \dataHours hours.

\subsection{Length and Structure}

\begin{figure}[t]
\centering
\includegraphics[width=0.82\textwidth]{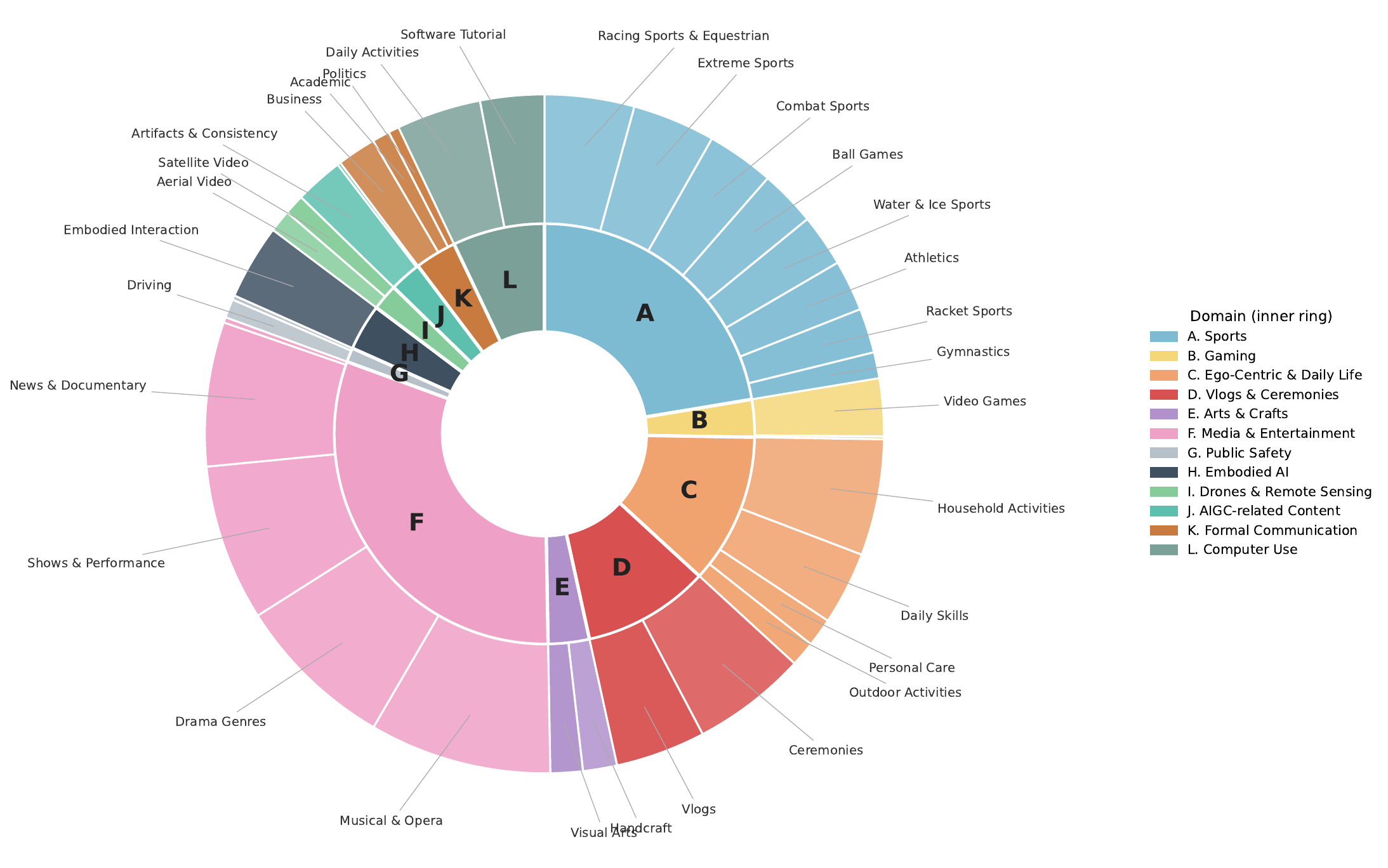}
\caption{\textbf{Two-ring content taxonomy of the \Kairos corpus.} Inner ring: the \videoDomains parent domains (codes A--L). Outer ring: the \videoCategories categories nested within domains, with leader-line labels. Wedge sizes are video counts. The corpus is non-uniform but well-spread: \texttt{A.~Sports} and \texttt{F.~Media \& Entertainment} together account for the largest share, while small-tail domains (\texttt{G.~Public Safety}, \texttt{H.~Embodied AI}, \texttt{I.~Drones \& Remote Sensing}) are kept at single-digit shares so that downstream evaluation can isolate domain-specific failures.}
\label{fig:dataset-pie}
\end{figure}

\begin{figure}[t]
\centering
\includegraphics[width=0.78\textwidth]{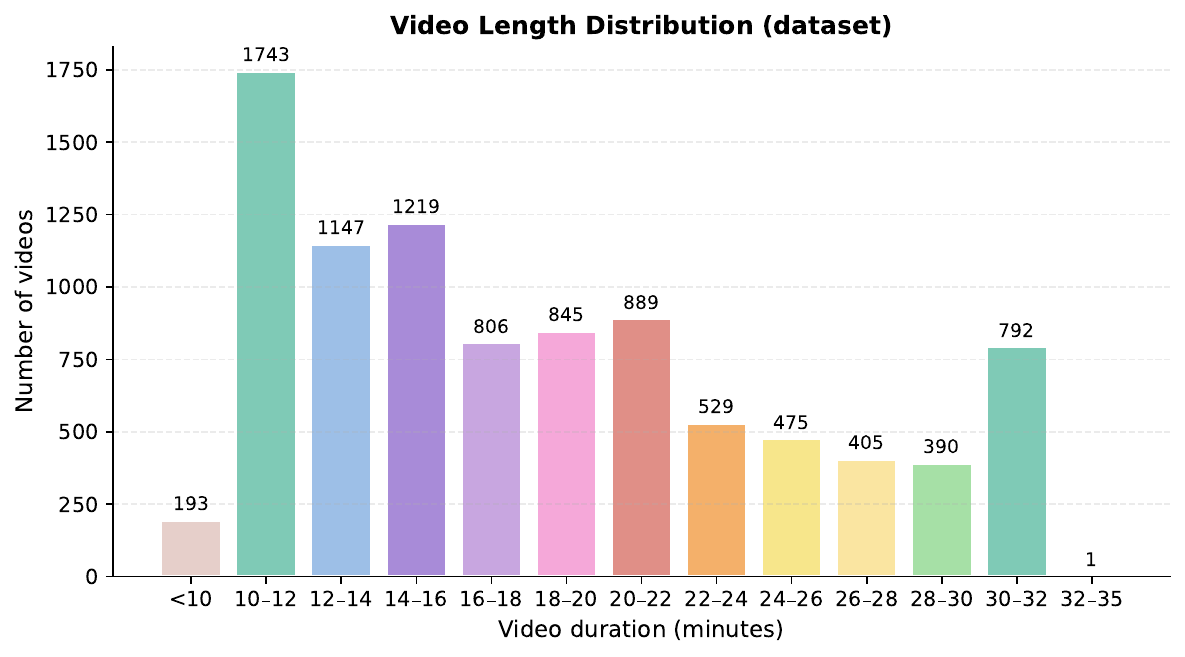}
\caption{\textbf{Per-video duration distribution} on the YouTube partition (\dataYouTubeVideos videos, \dataYouTubeHours hours). Median \dataDurationMedianMin\,min, mean \dataDurationMeanMin\,min, max \dataDurationMaxMin\,min. The bulk of the corpus sits in the \dataMinMinutes--\dataMaxMinutesEnd-min target band; the small \texttt{<10} bar reflects \texttt{\_partNN} tail fragments left after splitting longer sources rather than truncation.}
\label{fig:dataset-duration}
\end{figure}

\begin{figure}[t]
\centering
\includegraphics[width=0.78\textwidth]{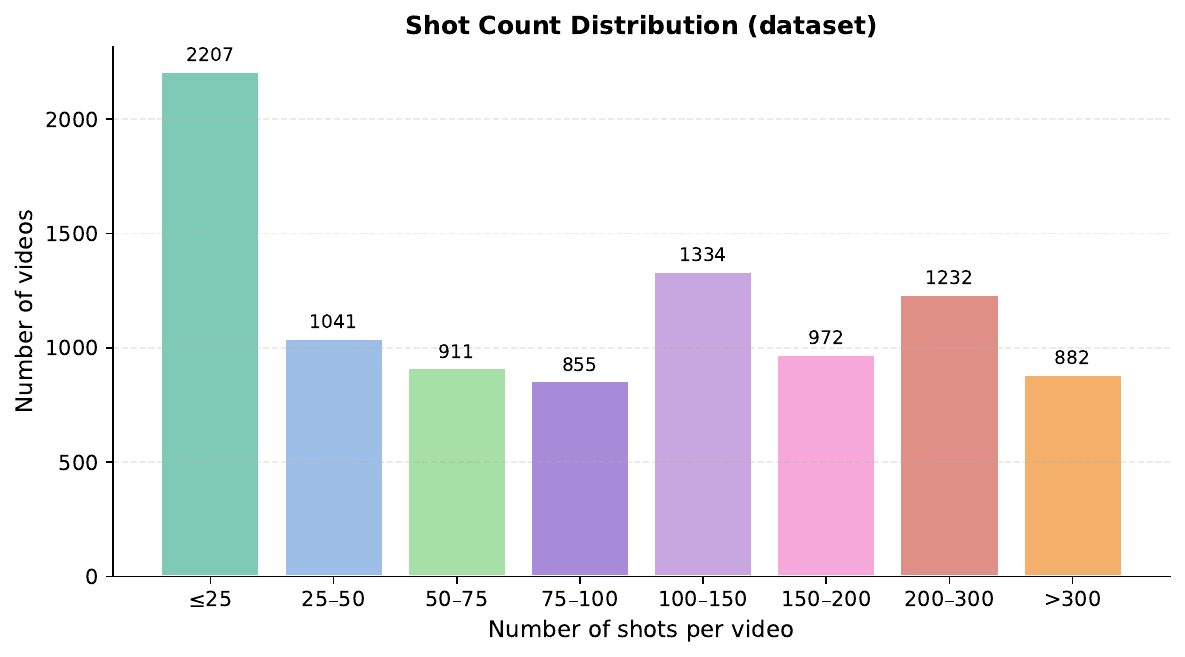}
\caption{\textbf{Shots-per-video distribution} on the YouTube partition. Median \dataShotsPerVideoMedian shots per video, mean \dataShotsPerVideoMean, max \dataShotsPerVideoMax, and \dataShotsTotal shots in total. The right tail is dominated by fast-cut sports broadcasts and clip compilations and motivates the differential-frame chain in the annotation pipeline.}
\label{fig:dataset-shot-count}
\end{figure}

\begin{figure}[t]
\centering
\includegraphics[width=0.78\textwidth]{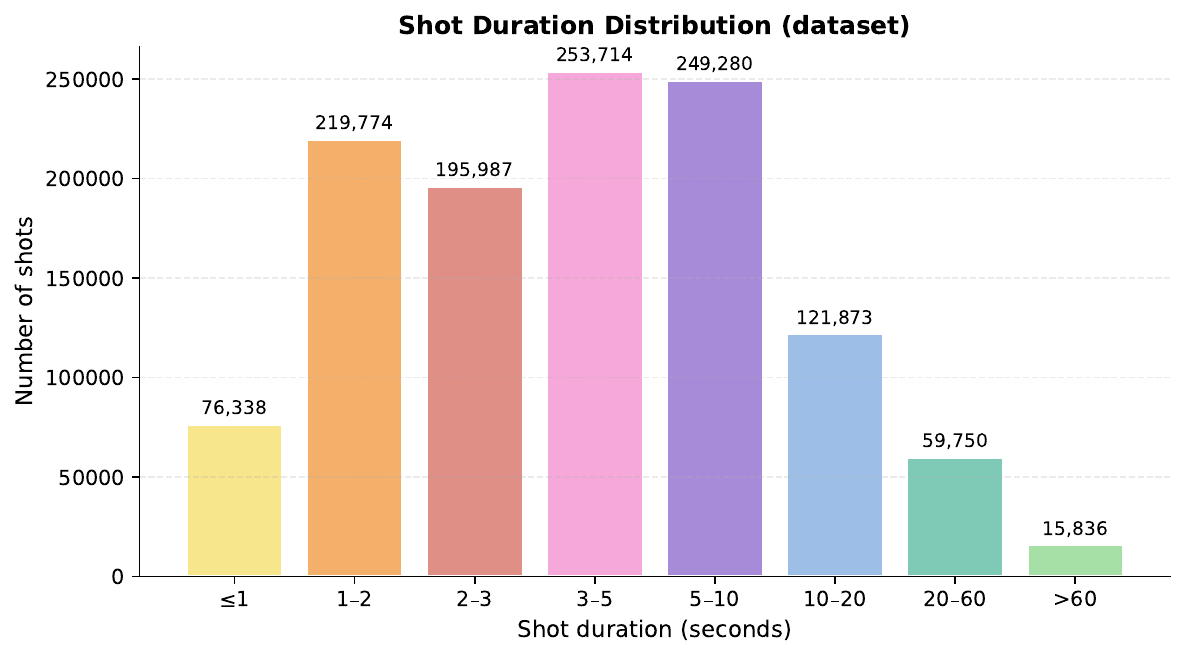}
\caption{\textbf{Per-shot duration distribution} on the YouTube partition. Median \dataShotDurMedianSec\,s, mean \dataShotDurMeanSec\,s. The right tail (held shots, $>$60\,s) is what makes Time-axis questions non-trivial in the derived benchmark, and is also where the \pipeReanchorInterval-frame re-anchor schedule is exercised.}
\label{fig:dataset-shot-dur}
\end{figure}

\begin{figure}[t]
\centering
\includegraphics[width=\textwidth]{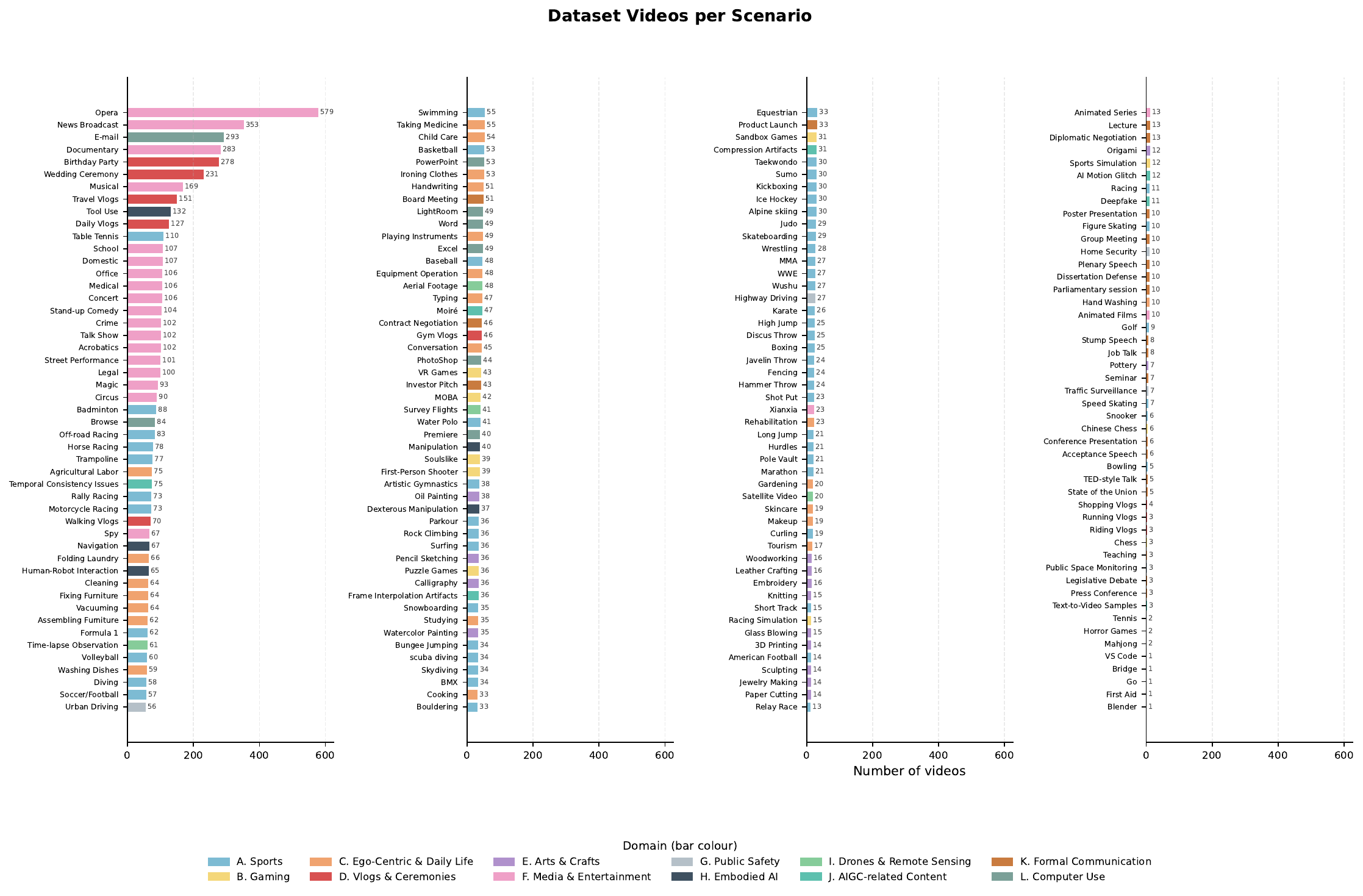}
\caption{\textbf{Per-scenario video count} across all \videoScenarios scenarios of the corpus, grouped by parent category and parent domain (bar color). Bars are sorted within each category by descending count; the resulting distribution is long-tailed but non-degenerate, with no single scenario dominating and every scenario receiving a meaningful share of the corpus.}
\label{fig:dataset-scenarios}
\end{figure}

The corpus's content taxonomy and length distribution are summarized in Figs.~\ref{fig:dataset-pie}--\ref{fig:dataset-shot-dur}, with the deepest per-scenario decomposition in Fig.~\ref{fig:dataset-scenarios}.
Together these substantiate the design claim that \Kairos lives in the \emph{multi-shot, long-context} regime that single-clip benchmarks do not exercise.

\section{Benchmark Construction Details}
\label{app:benchmark}

\subsection{Source-type Samplers and Tier Mapping}

Each of the \benchSourceTypes source types corresponds to a single sampler over the timestamped annotation stream.
A sampler returns the materials needed for question generation --- the relevant annotations, their timestamps, the evidence span, neighbouring reference-frame context, and a pool of candidate distractors --- without ever accessing the raw frames.
The temporal tier of a generated question is determined by the realized evidence span itself, not assigned post hoc by the LLM.
Each sampler also restricts the capability labels that the generator may emit (e.g.\ \texttt{ref\_audio} can only be tagged with \texttt{A5\_audio\_comprehension}, while \texttt{transition} is tagged with \texttt{C2\_narrative\_transition}).
Table~\ref{tab:source-quotas} reports the per-source-type composition of the released benchmark, alongside the tier range that each sampler's evidence span can fall into.

\begin{table}[t]
\centering
\small
\caption{\textbf{Per-source-type composition of \KairosBench} after audit and human review. \emph{Allowed tiers} is the range of evidence spans the sampler can realise; the tier of a question is set by the realized span, not by the source type.}
\label{tab:source-quotas}
\begin{tabular}{@{}llcr@{}}
\toprule
\textbf{Source type} & \textbf{Evidence shape} & \textbf{Allowed tiers} & \textbf{\#Q} \\
\midrule
\texttt{ref\_perception}   & single reference frame, scene + entities + relations & T1 & \srcRefPerception \\
\texttt{ref\_ocr}          & single reference frame, on-screen text                & T1 & \srcRefOcr \\
\texttt{ref\_audio}        & single reference frame, aligned ASR / environment     & T1 & \srcRefAudio \\
\texttt{diff\_change}      & ref + several differential frames inside one shot     & T2 & \srcDiffChange \\
\texttt{diff\_sequence}    & full differential chain of one shot                   & T2 & \srcDiffSequence \\
\texttt{entity\_tracking}  & cross-shot entity reappearance                        & T3--T5 & \srcEntityTracking \\
\texttt{transition}        & two boundary frames + adjacent shot descriptions      & T2--T3 & \srcTransition \\
\texttt{cross\_shot}       & several adjacent or near-adjacent shots               & T3--T4 & \srcCrossShot \\
\texttt{long\_range}       & evidence span $>\!300$\,s                             & T4--T5 & \srcLongRange \\
\texttt{full\_video}       & whole-video integration                               & T5 & \srcFullVideo \\
\bottomrule
\end{tabular}
\end{table}

\noindent\textbf{Atomic generation.}
Given the sampled evidence, the neighbouring reference-frame context before and after the target evidence, and a coarse temporal hint, \pipeGen returns a structured JSON object containing four fields: \texttt{question}, \texttt{answer}, \texttt{distractors}, and \texttt{reasoning}.
All four are produced atomically in the same call.
This encourages internal consistency between the answer and the reasoning, and avoids a separate rewriting stage that could make the distractors superficially different from the correct answer.

\subsection{Cross-benchmark Text-only Leakage --- Protocol}
\label{app:leakage}

The cross-benchmark leakage probe in Table~\ref{tab:cross-benchmark-leakage} of the main text fixes the solver, the prompt, the option-shuffle protocol, and the sample size identically across all benchmarks:
\begin{itemize}[leftmargin=*, labelsep=0.4em, itemsep=0pt, topsep=0pt, partopsep=0pt, parsep=0pt]
    \item \textbf{Solver.} \leakageSolver, no video and no audio, $K{=}1$ shuffle, seed 42, greedy decoding, with the smallest thinking budget the model accepts (128 tokens).
    \item \textbf{Sample size.} \reviewLeakageProbe questions per benchmark, stratified over capability/tier/source where applicable, otherwise uniform.
    \item \textbf{Prompt.} A fixed system prompt states that no video or images are available and asks for the answer letter on the first line followed by one sentence of rationale on the second; the user message contains the question stem and the shuffled options. No in-context examples and no reasoning before the answer.
    \item \textbf{Random baseline.} Computed per benchmark from the realized option counts ($1/N$ for fixed-$N$ benchmarks, and the per-question mean of $1/N_i$ for variable-option benchmarks).
    \item \textbf{Leakage.} Solver accuracy minus random baseline, reported in percentage points.
\end{itemize}

\begin{figure}[t]
\centering
\begin{minipage}[t]{0.49\textwidth}
\centering
\includegraphics[width=\linewidth]{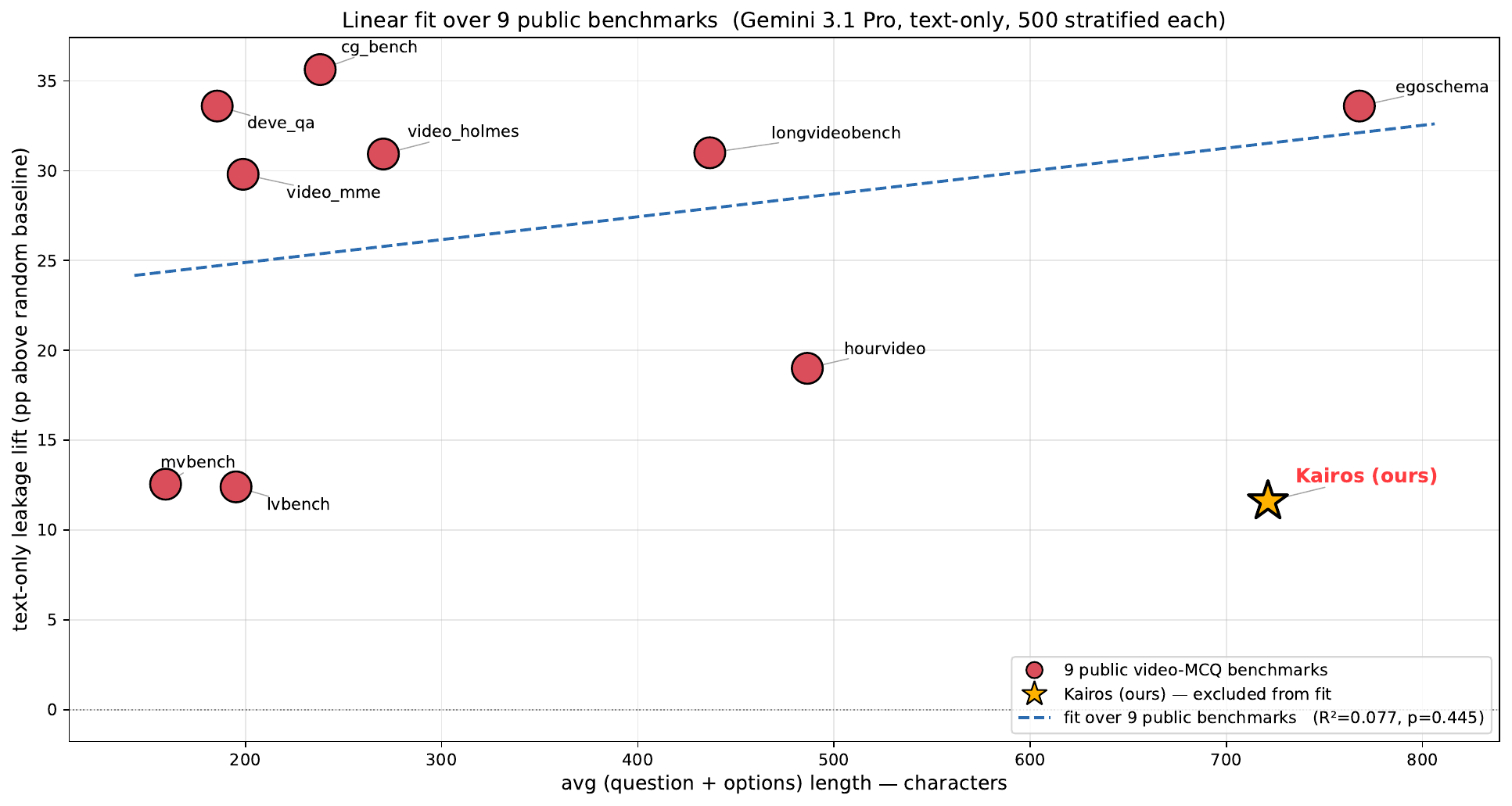}
\caption{\textbf{Leakage vs.\ average MCQ length.} Linear fit of text-only-leakage (pp) on average total characters per question across the ten benchmarks of Table~\ref{tab:cross-benchmark-leakage} of the main text. \KairosBench is to the right of the cloud (long stems) and below the line (lower leakage than length predicts).}
\label{fig:leakage-length-fit}
\end{minipage}
\hfill
\begin{minipage}[t]{0.49\textwidth}
\centering
\includegraphics[width=\linewidth]{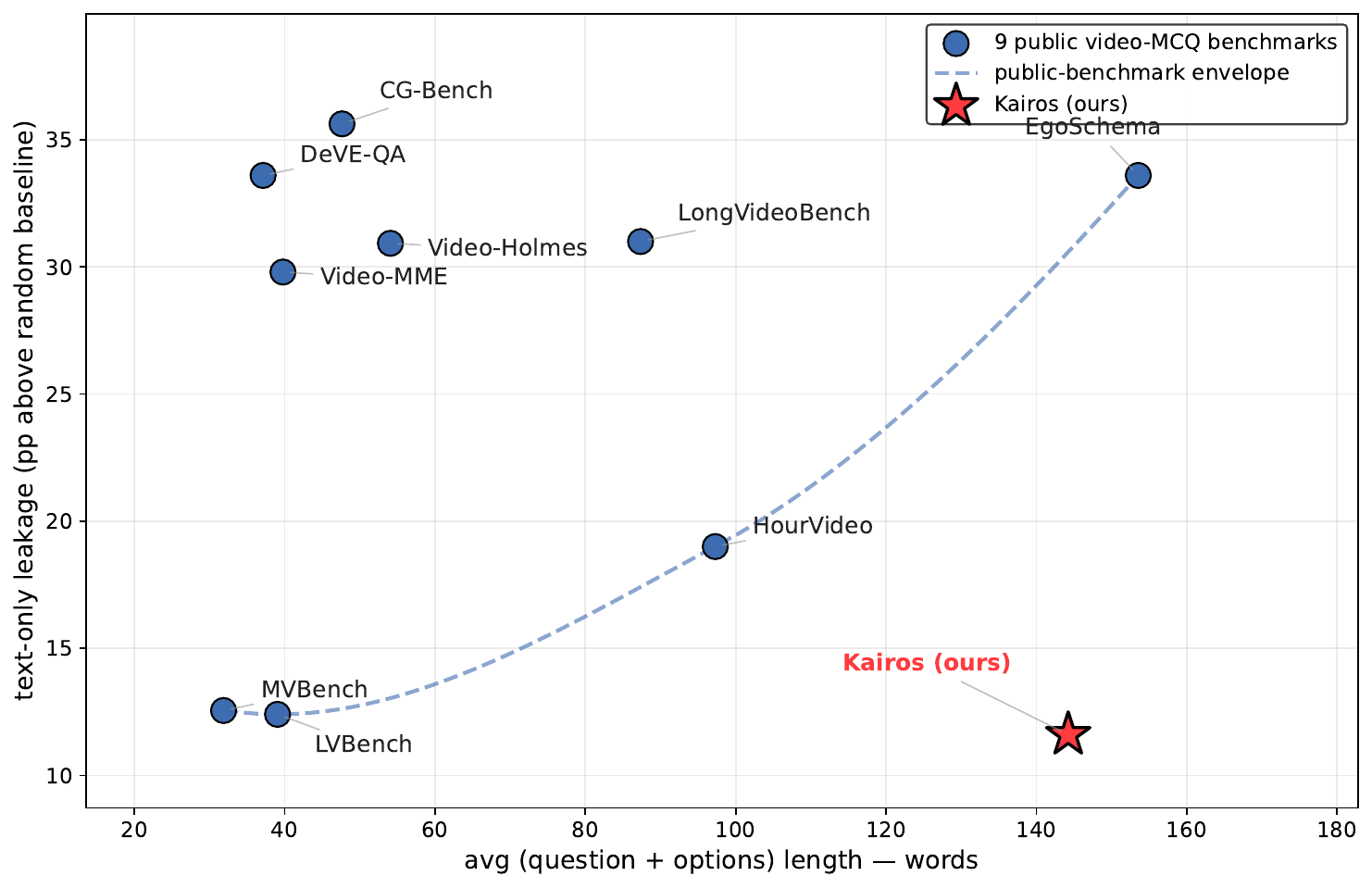}
\caption{\textbf{Leakage--length scatter.} Per-benchmark scatter underlying the fit in Fig.~\ref{fig:leakage-length-fit}. Each dot is one of the ten benchmarks; axes are average total MCQ characters and the absolute lift over the random baseline. \Kairos is the rightmost low-leakage point.}
\label{fig:leakage-length-scatter}
\end{minipage}
\end{figure}

Figure~\ref{fig:leakage-length-fit} fits a linear regression of leakage on average total characters across the ten benchmarks; \KairosBench has the longest stems among the four-option benchmarks but lies \emph{below} the regression line, i.e.\ its leakage is lower than its length would predict.
Figure~\ref{fig:leakage-length-scatter} shows the underlying scatter, broken down by benchmark.

\subsection{Benchmark Distribution Diagnostics}

\begin{figure}[t]
\centering
\includegraphics[width=0.78\textwidth]{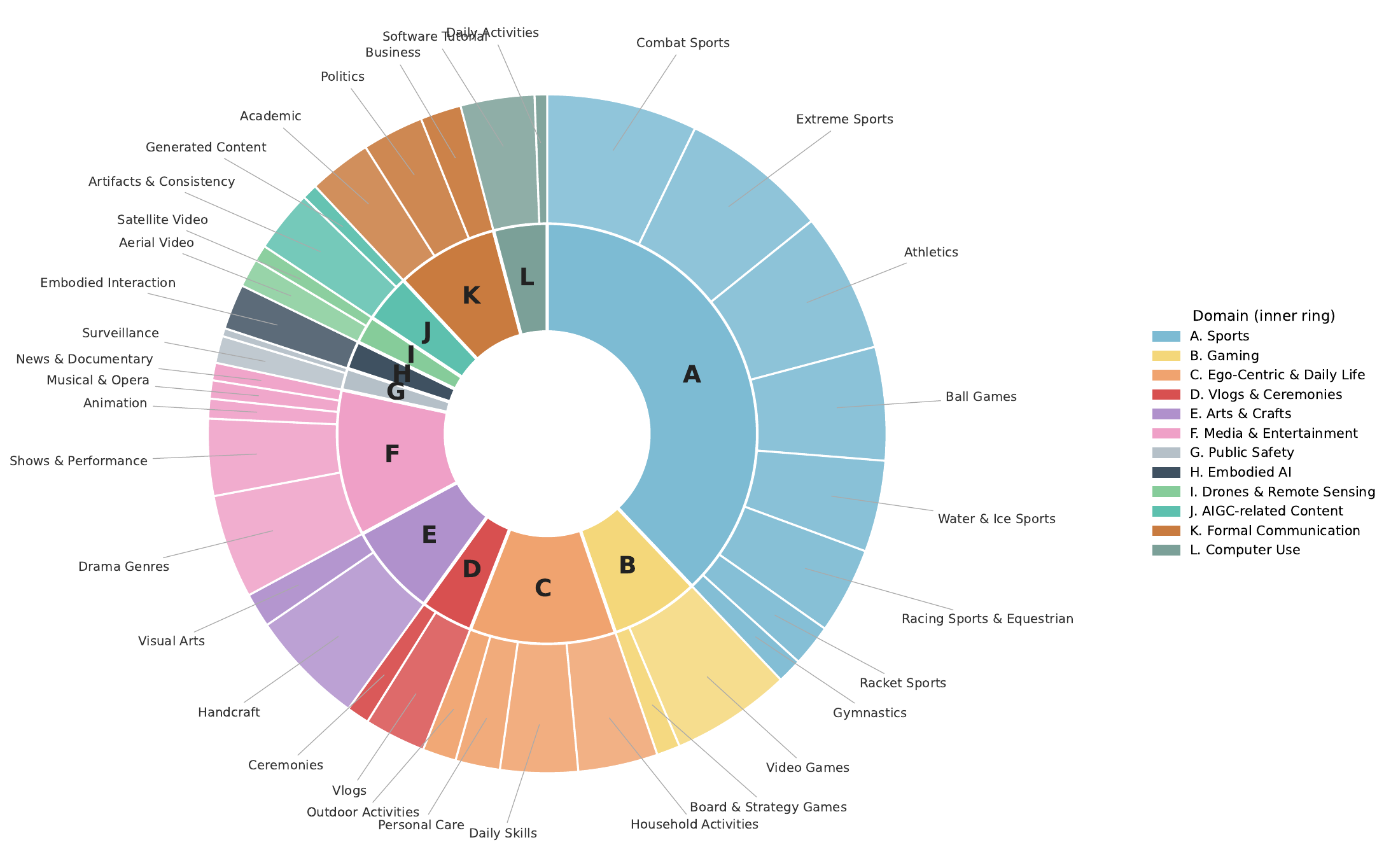}
\caption{\textbf{Two-ring view of the capability axis of \KairosBenchEnd.} Inner ring: the four cognitive levels A (Perception, Space), B (Events, within-shot Dynamics), C (Temporal, cross-shot Dynamics), D (Localisation, holistic). Outer ring: the \benchCapabilities capability cells nested within each level.}
\label{fig:bench-pie}
\end{figure}

\begin{figure}[t]
\centering
\includegraphics[width=\textwidth]{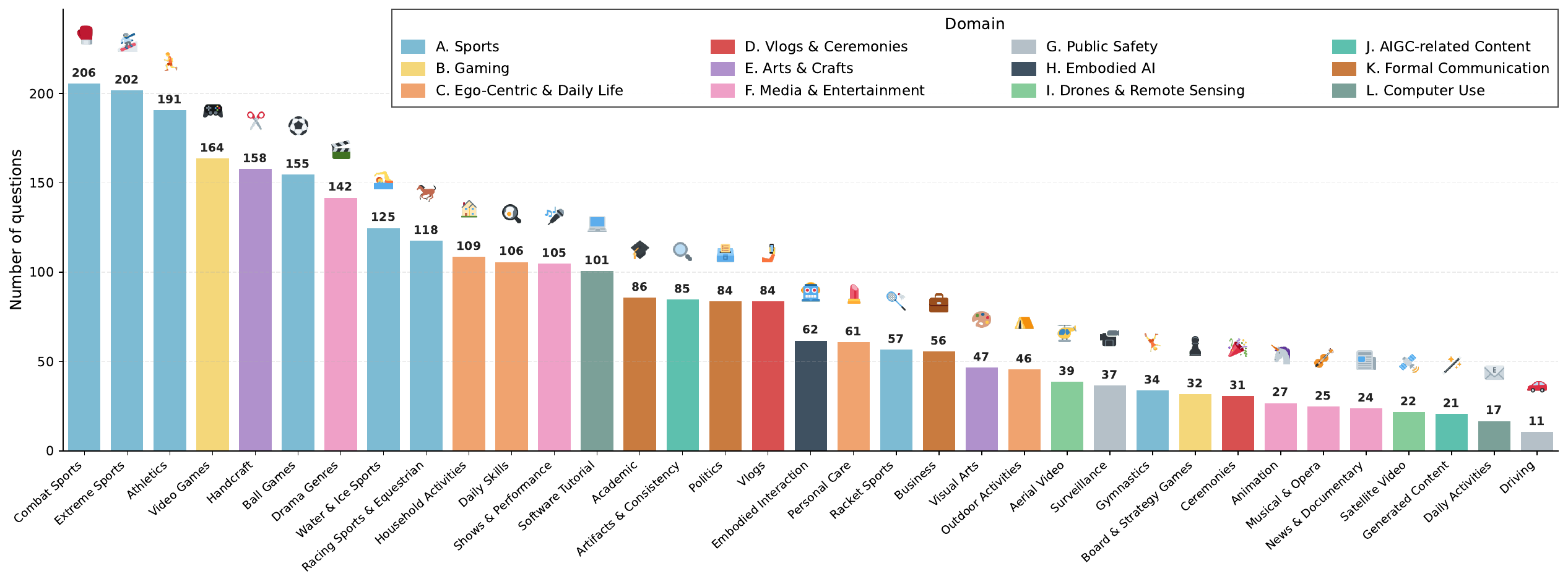}
\caption{\textbf{Per-category video count of \KairosBench} (\benchVideos videos), grouped by parent domain (bar color). The benchmark preserves the long-tail shape of the corpus (Fig.~\ref{fig:dataset_categories} of the main text); no domain is dropped, and the small-tail domains \texttt{G}, \texttt{H}, \texttt{I} are over-sampled relative to their share of the corpus to keep per-domain question counts non-trivial.}
\label{fig:bench-categories}
\end{figure}

\begin{figure}[t]
\centering
\includegraphics[width=\textwidth]{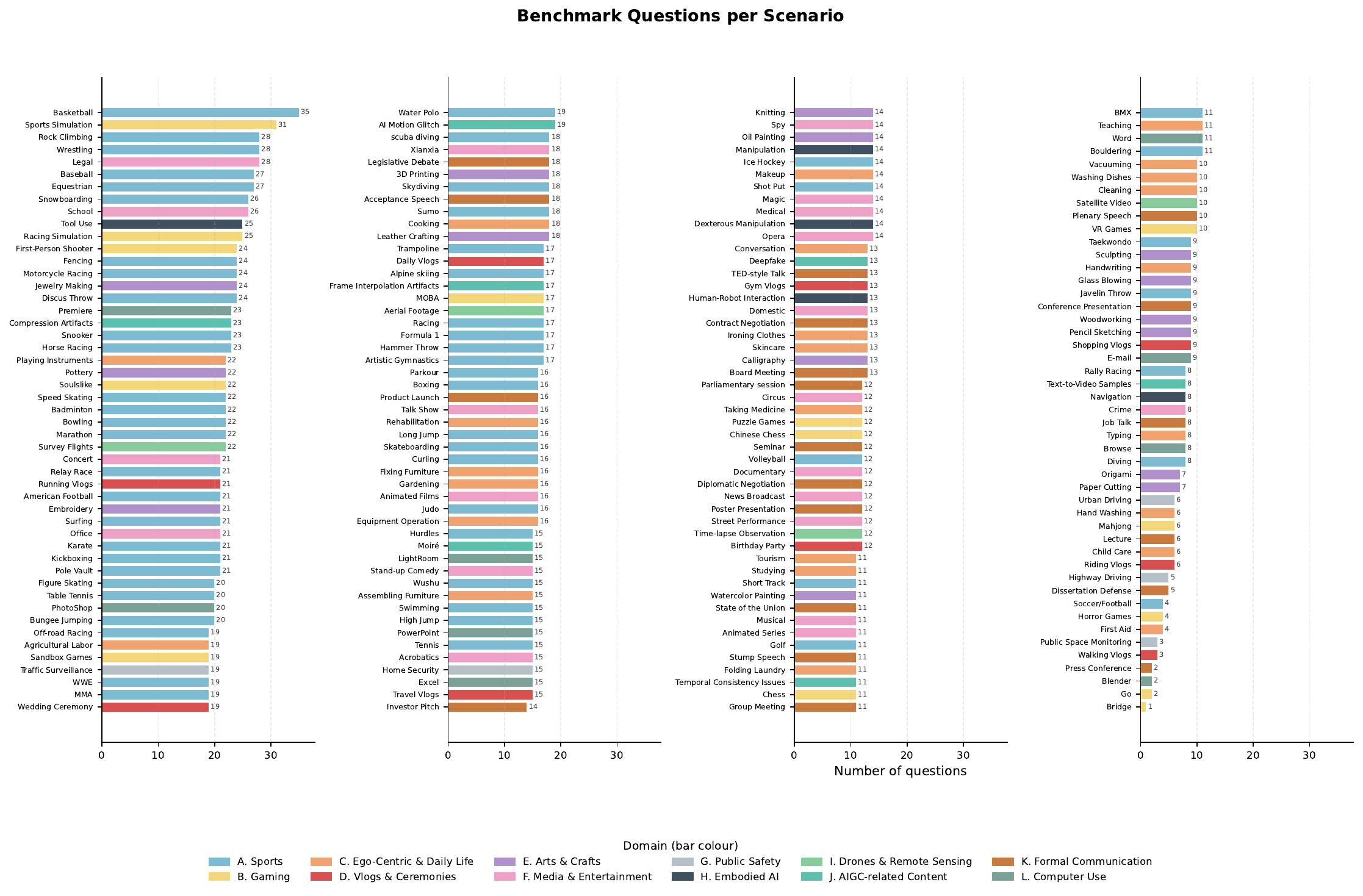}
\caption{\textbf{Per-scenario video count of \KairosBench} across the \videoScenarios scenarios of the content taxonomy. Bars are colored by parent domain. The benchmark video pool is intentionally spread thin across scenarios rather than concentrated in a few high-volume cells, so that per-scenario evaluation is well-defined for as many cells as possible at the cost of small absolute counts in the long tail.}
\label{fig:bench-scenarios}
\end{figure}

\begin{figure}[t]
\centering
\begin{minipage}[t]{0.32\textwidth}
\centering
\includegraphics[width=\linewidth]{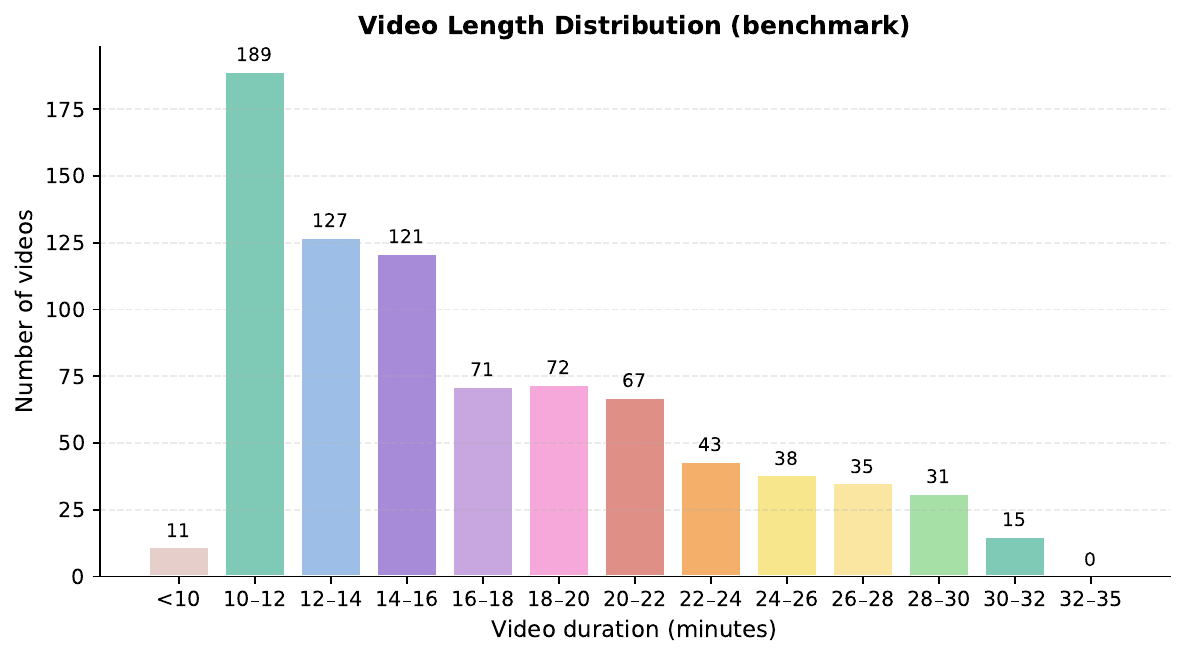}
\caption{\textbf{Per-video duration} on the \benchVideos benchmark videos.}
\label{fig:bench-duration}
\end{minipage}
\hfill
\begin{minipage}[t]{0.32\textwidth}
\centering
\includegraphics[width=\linewidth]{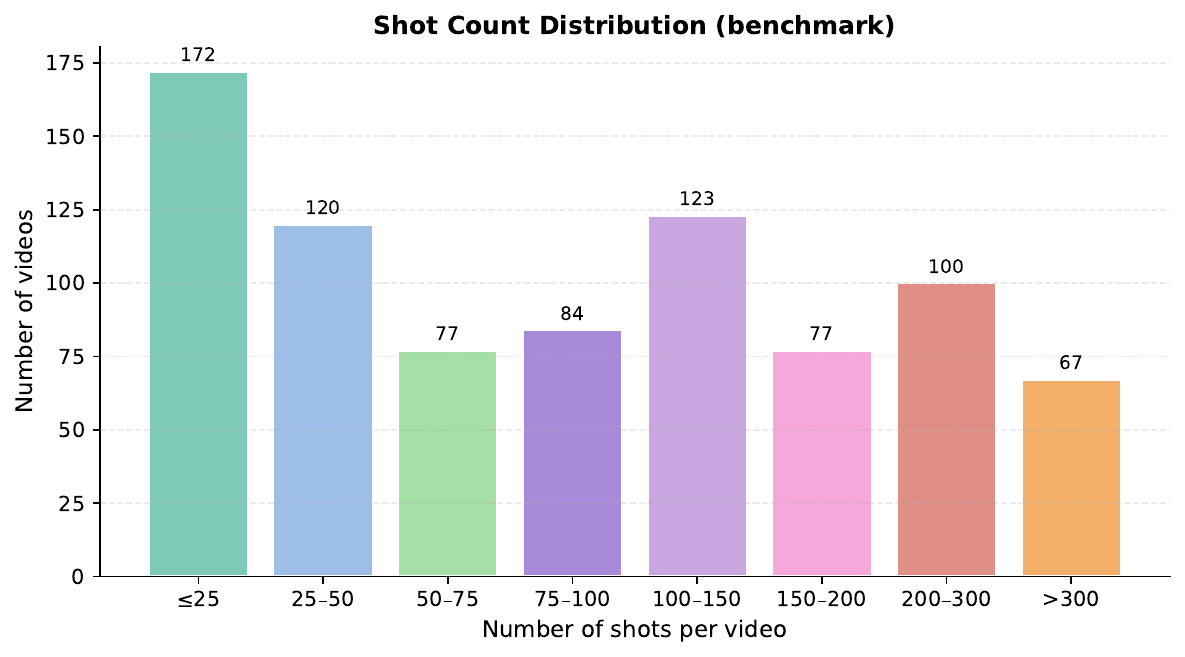}
\caption{\textbf{Shots per benchmark video.}}
\label{fig:bench-shot-count}
\end{minipage}
\hfill
\begin{minipage}[t]{0.32\textwidth}
\centering
\includegraphics[width=\linewidth]{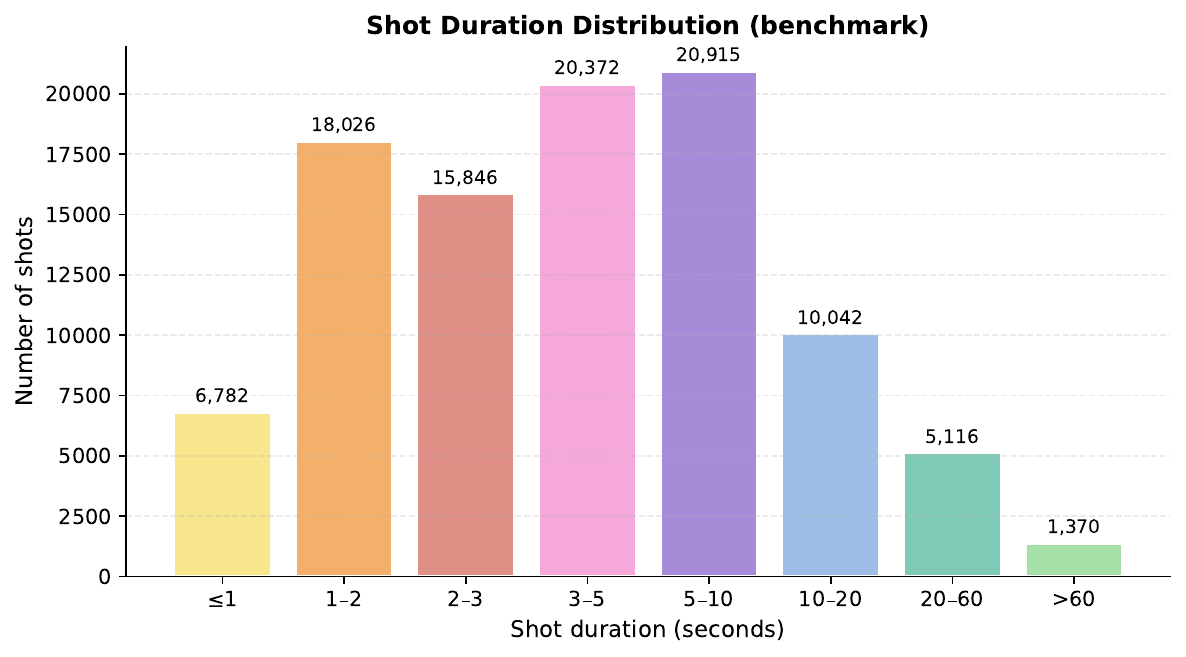}
\caption{\textbf{Per-shot duration} on the benchmark videos.}
\label{fig:bench-shot-dur}
\end{minipage}
\end{figure}

Figures~\ref{fig:bench-pie}--\ref{fig:bench-shot-dur} expand the three-axis summary in Fig.~\ref{fig:benchmark_statistics} of the main text to the content and structural axes, none of which are exposed in the main paper.
Figure~\ref{fig:bench-pie} reframes the capability axis as a two-ring pie (axis $\to$ capability) so that the relative weight of holistic Axis-D cells is legible.
Figure~\ref{fig:bench-categories} reports the benchmark video count by content category, and Fig.~\ref{fig:bench-scenarios} the per-scenario decomposition; both confirm that the \benchVideos benchmark videos preserve the corpus-level taxonomy distribution rather than over-sampling a small subset.
Figures~\ref{fig:bench-duration}--\ref{fig:bench-shot-dur} report per-video duration, shots per video, and per-shot duration on the benchmark subset.

\section{Evaluation Protocol}
\label{app:eval}

We evaluate \evalModels models in total: \evalClosedModels closed-source proprietary and \evalOpenModels open-weight.
A unified runner groups questions by \texttt{video\_id}, prepares the video once per video for the relevant backend, then dispatches every question for that video in parallel (API: ThreadPoolExecutor) or batched (vLLM: \texttt{generate\_batch}).

\subsection{Backends and Video-input Negotiation}

\noindent\textbf{Five backends.}
The runner instantiates one of five backends per model based on its registry entry:
\begin{itemize}[leftmargin=*, labelsep=0.4em, itemsep=0pt, topsep=0pt, partopsep=0pt, parsep=0pt]
    \item \textbf{vLLM} (open-weight): frame mode for every family except \textsc{llava-video-7b}, which is fed the video directly.
    \item \textbf{HuggingFace Transformers} (open-weight): used for the few open-weight models without a vLLM video implementation, in frame mode only.
    \item \textbf{OpenAI direct} (closed-source + OpenRouter-proxied open-weight): probes \texttt{data:video/mp4} native video once per model and caches the result; falls back to $N$ uniformly subsampled frames otherwise.
    \item \textbf{Gemini direct}: File API upload + 2\,s processing poll, cached per video path. Native video is the only input mode.
    \item \textbf{Anthropic direct}: frame mode only (no native video support in the public API).
\end{itemize}

\noindent\textbf{Native-video models.}
Only \evalNativeVideoModels run with native video input in our evaluation; every other model receives uniformly subsampled frames at the per-model budget reported in Table~\ref{tab:eval-config}.

\subsection{Per-model Decoding and Frame Budget}

Table~\ref{tab:eval-config} enumerates the eval-time configuration used for every model in the leaderboard of Table~\ref{tab:main_results} of the main text.
The frame budget is the maximum number of frames the runner sends per question; the actual count for a given question equals $\min(\textit{budget}, \lfloor \textit{video\_dur} \cdot \pipeFps \rfloor)$ so that very short videos are never up-sampled past their native \pipeFps FPS rate.
Decoding is greedy (temperature $0$) for every model except \textsc{gpt-5.5}, whose API only accepts its default sampling temperature; for the OpenQA run we keep the same decoding so that the only intentional change between MCQ and OpenQA is the answer format.

\begin{table}[t]
\centering
\small
\caption{\textbf{Per-model evaluation configuration.} \emph{Type} is the backend used. \emph{TP} is the tensor-parallel size on H200 GPUs (vLLM only). \emph{Frames} is the per-question frame budget (\emph{native} = native video input). \emph{Max tokens} is the decoding cap. Sampling is greedy throughout.}
\label{tab:eval-config}
\setlength{\tabcolsep}{4pt}
\begin{tabular}{@{}llrlrl@{}}
\toprule
\textbf{Model} & \textbf{Type} & \textbf{TP} & \textbf{Frames} & \textbf{Max tokens} & \textbf{Notes} \\
\midrule
\multicolumn{6}{l}{\textcolor{gray}{\textit{Closed-source proprietary}}} \\
\textsc{gemini-3.1-pro}      & gemini    & --- & native & 1024 & video-native \\
\textsc{gemini-2.5-flash}    & gemini    & --- & native & 512  & video-native \\
\textsc{gpt-5.5}             & openai    & --- & 256    & 2048 & forced frame mode \\
\textsc{gpt-4o}              & openai    & --- & 64     & 1024 & forced frame mode \\
\textsc{gpt-4o-mini}         & openai    & --- & 128    & 1024 & forced frame mode \\
\textsc{nova-2-lite}         & openai    & --- & 32     & 1024 & forced frame mode \\
\textsc{seed-2.0-lite}       & openai    & --- & 32     & 1024 & forced frame mode \\
\midrule
\multicolumn{6}{l}{\textcolor{gray}{\textit{Open-weight}}} \\
\textsc{internvl3-78b}       & vLLM      & 4   & 16     & 512  & frame mode \\
\textsc{internvl3.5-38b}     & vLLM      & 2   & 12     & 512  & frame mode \\
\textsc{internvl3-8b}        & vLLM      & 2   & 16     & 512  & frame mode \\
\textsc{internvl3.5-8b}      & vLLM      & 2   & 12     & 2048 & structured-output letter \\
\textsc{glm-4.5v}            & vLLM      & 4   & 16     & 64   & \texttt{enable\_thinking=False} \\
\textsc{glm-4v-9b}           & HF        & --- & 1      & 512  & 1-frame ceiling per the model card \\
\textsc{step3-vl-10b}        & vLLM      & 2   & 16     & 8192 & frame mode \\
\textsc{qwen3-vl-30b-a3b}    & vLLM      & 2   & 16     & 512  & frame mode \\
\textsc{qwen3-vl-8b}         & vLLM      & 1   & 16     & 512  & frame mode \\
\textsc{qwen2.5-vl-7b}       & vLLM      & 1   & 16     & 512  & frame mode \\
\textsc{mimo-vl-7b}          & vLLM      & 1   & 16     & 4096 & structured-output letter \\
\textsc{cogvlm2-video-13b}   & HF        & --- & 16     & 1024 & frame mode \\
\textsc{gemma-4-31b}         & openai    & --- & 32     & 1024 & forced frame mode \\
\textsc{llava-video-7b}      & vLLM      & 1   & native & 512  & video-native; structured-output letter \\
\bottomrule
\end{tabular}
\end{table}

\noindent\textbf{Scoring.}
A model output is parsed by first stripping any \texttt{<think>} block, then taking the last explicit \texttt{Answer:X} if present, otherwise a leading bracketed or bare letter; outputs with no such letter are counted as incorrect.
Per-tier, per-capability, per-source, and per-domain accuracies are computed by simple bucket-then-average over the parsed letters.
Three open-weight models (\textsc{internvl3.5-8b}, \textsc{mimo-vl-7b}, \textsc{llava-video-7b}) are decoded with vLLM \texttt{structured\_outputs} that constrain the final token to a single letter, because their unconstrained outputs occasionally trail off into reasoning without committing to a choice.

\subsection{Open-ended QA Judge}

\noindent\textbf{Why an OpenQA pass.}
A multiple-choice task is easy to score but lossy: a model can recognise the right answer without being able to produce it.
We therefore additionally run an OpenQA pass on the same \benchReviewedMCQs questions, asking each model to generate a free-form answer rather than pick a letter.
Decoding stays greedy; the per-model token budget is raised to at least 256 tokens so that free-form answers are not truncated.

\noindent\textbf{Judge.}
Each generation is scored against the reference answer by \openqaJudge, which sees only the question stem, the reference answer, and the candidate; no video, no audio, no chain-of-thought.
The judge returns an integer 0--3 in the MMBench-Video style, and we report binarised accuracy at threshold \openqaThreshold.
The leaderboard is reproduced in Table~\ref{tab:openqa_leaderboard} of the main text.

\section{Fine-tuning Recipe}
\label{app:training}

\subsection{Hyperparameters}

Table~\ref{tab:ft-hparams} summarizes every fine-tuning hyperparameter.
The recipe deliberately stays close to a vanilla LLaMA-Factory \texttt{qwen2\_5\_vl} template: LoRA on the seven attention/MLP projections, frozen vision tower, single epoch, cosine schedule.
The only non-default choice is the cutoff length, which is raised to \ftCutoffLen tokens to fit the long anchor-based prompts at \ftFrames frames.

\begin{table}[t]
\centering
\small
\caption{\textbf{Fine-tuning hyperparameters.} The recipe uses LLaMA-Factory's \texttt{qwen2\_5\_vl} template; the seven LoRA target modules are the standard transformer projections.}
\label{tab:ft-hparams}
\begin{tabular}{@{}ll@{}}
\toprule
\textbf{Setting} & \textbf{Value} \\
\midrule
\multicolumn{2}{l}{\textcolor{gray}{\textit{Model and parameterisation}}} \\
Base model                 & \ftBaseModel \\
Framework                  & \ftFrameworkFT \\
Adapter                    & LoRA, rank \ftLoraRank, $\alpha=$\ftLoraAlpha \\
LoRA target modules        & \texttt{q,k,v,o,gate,up,down}\_proj \\
Frozen modules             & vision tower (multi-modal projector trainable) \\
Precision                  & \ftPrecisionFT \\
Distributed                & \ftDDP \\
Hardware                   & \ftHardwareFT \\
\midrule
\multicolumn{2}{l}{\textcolor{gray}{\textit{Optimization}}} \\
Optimizer                  & AdamW (LLaMA-Factory default) \\
Learning rate              & \ftLR \\
Schedule                   & cosine, warmup ratio \ftWarmup \\
Weight decay               & 0.01 \\
Per-device batch size      & 1 \\
Gradient accumulation      & 8 \\
Effective batch size       & \ftEffectiveBatch \\
Epochs                     & \ftEpochsFT \\
Cutoff length              & \ftCutoffLen tokens \\
\midrule
\multicolumn{2}{l}{\textcolor{gray}{\textit{Visual input}}} \\
Frame budget per sample    & \ftFrames \\
Sampler                    & uniform \\
Input resolution           & $420\!\times\!420$ pixels per frame (image and video pixels) \\
Frame source               & pre-extracted \pipeFps FPS JPEGs (no re-decoding) \\
\bottomrule
\end{tabular}
\end{table}

\noindent\textbf{Eval-time frame ablation.}
At eval time, the fine-tuned checkpoint is decoded with $\{16, 32, 64\}$ uniformly subsampled frames per question; the headline numbers across these frame budgets and across the four evaluation benchmarks are reported in Table~\ref{tab:training_qwen}.
%

\section{Limitations and Broader Impact}
\label{app:limits}

\subsection{Limitations}

\noindent\textbf{VLM bias.}
All annotations are produced by \texttt{Qwen3-VL-8B-Instruct}~\citep{bai2025qwen3}, so a systematic perception bias of that VLM also biases \Kairos descriptions; running the pipeline with a different VLM is straightforward and would help quantify the bias.

\noindent\textbf{Language-only entity matching.}
Cross-shot entity matching is language-only, so for visually-hard-to-describe entities (e.g.\ visually similar but semantically distinct people in crowd scenes), label-exact matching splits an identity whenever the model paraphrases an established name instead of reusing it, and merges two mentions only when it assigns them the same label.
This is addressable by adding a vision-only re-identification signal as a visual matching stage, which we leave to future work to keep the pipeline VLM-only.

\noindent\textbf{Residual leakage.}
Even after the multi-model audit and human review, the cross-benchmark leakage probe still finds \leakageKairos lift over random under \leakageSolver --- the lowest of the ten benchmarks tested but not zero.
We attribute this residual to the linguistic anchors that the questions include in lieu of numeric timestamps; removing those anchors would reduce leakage further but would also degrade question grounding for non-visual solvers, so we leave the trade-off explicit rather than tune it.

\noindent\textbf{Per-tier population imbalance.}
Two cells of the capability axis have small populations (\texttt{D1\_narrative\_summarisation} 5 questions, \texttt{C2\_narrative\_transition} 14 questions).
Per-cell accuracy on those cells is therefore high-variance and should be read as a coarse signal rather than a precise number.
The headline overall accuracy and the per-tier breakdown are unaffected.

\subsection{Broader Impact}

\noindent\textbf{Intended uses.}
\Kairos is intended for the development and evaluation of long-form video understanding systems in three modes: (i) as supervision data for video-language pretraining and fine-tuning; (ii) as a benchmark for long-form video QA; (iii) as a substrate for evaluating video-language alignment, entity tracking, and temporal grounding methods that go beyond clip-level decisions.

\noindent\textbf{Risks.}
The dataset contains only annotations of public, platform-distributed videos, with platform-takedown semantics preserved.
Personally-identifying information that appears in the source videos (e.g.\ recognisable individuals in sports broadcasts) is not added by the annotation pipeline; faces and identities are referenced only at the level the source already exposes them.
Misuse risks are those typical of any large video corpus; the annotation-side frames are not redistributed, which limits this surface.


%% file: iclr2027_conference.bib
@article{laptev2005space,
  title={On space-time interest points},
  author={Laptev, Ivan},
  journal={International journal of computer vision},
  volume={64},
  number={2},
  pages={107--123},
  year={2005},
  publisher={Springer}
}

@inproceedings{klaser2008spatio,
  title={A spatio-temporal descriptor based on 3d-gradients},
  author={Klaser, Alexander and Marsza{\l}ek, Marcin and Schmid, Cordelia},
  booktitle={BMVC 2008-19th British machine vision conference},
  pages={275--1},
  year={2008},
  organization={British Machine Vision Association}
}

@inproceedings{wang2013action,
  title={Action recognition with improved trajectories},
  author={Wang, Heng and Schmid, Cordelia},
  booktitle={Proceedings of the IEEE international conference on computer vision},
  pages={3551--3558},
  year={2013}
}

@article{wang2013dense,
  title={Dense trajectories and motion boundary descriptors for action recognition},
  author={Wang, Heng and Kl{\"a}ser, Alexander and Schmid, Cordelia and Liu, Cheng-Lin},
  journal={International journal of computer vision},
  volume={103},
  number={1},
  pages={60--79},
  year={2013},
  publisher={Springer}
}

@inproceedings{tran2015learning,
  title={Learning spatiotemporal features with 3d convolutional networks},
  author={Tran, Du and Bourdev, Lubomir and Fergus, Rob and Torresani, Lorenzo and Paluri, Manohar},
  booktitle={Proceedings of the IEEE international conference on computer vision},
  pages={4489--4497},
  year={2015}
}

@inproceedings{carreira2017quo,
  title={Quo vadis, action recognition? a new model and the kinetics dataset},
  author={Carreira, Joao and Zisserman, Andrew},
  booktitle={proceedings of the IEEE Conference on Computer Vision and Pattern Recognition},
  pages={6299--6308},
  year={2017}
}

@inproceedings{feichtenhofer2019slowfast,
  title={Slowfast networks for video recognition},
  author={Feichtenhofer, Christoph and Fan, Haoqi and Malik, Jitendra and He, Kaiming},
  booktitle={Proceedings of the IEEE/CVF international conference on computer vision},
  pages={6202--6211},
  year={2019}
}

@inproceedings{wang2016temporal,
  title={Temporal segment networks: Towards good practices for deep action recognition},
  author={Wang, Limin and Xiong, Yuanjun and Wang, Zhe and Qiao, Yu and Lin, Dahua and Tang, Xiaoou and Van Gool, Luc},
  booktitle={European conference on computer vision},
  pages={20--36},
  year={2016},
  organization={Springer}
}

@inproceedings{zhou2018temporal,
  title={Temporal relational reasoning in videos},
  author={Zhou, Bolei and Andonian, Alex and Oliva, Aude and Torralba, Antonio},
  booktitle={Proceedings of the European conference on computer vision (ECCV)},
  pages={803--818},
  year={2018}
}

@inproceedings{wang2018non,
  title={Non-local neural networks},
  author={Wang, Xiaolong and Girshick, Ross and Gupta, Abhinav and He, Kaiming},
  booktitle={Proceedings of the IEEE conference on computer vision and pattern recognition},
  pages={7794--7803},
  year={2018}
}

@inproceedings{bertasius2021space,
  title={Is space-time attention all you need for video understanding?},
  author={Bertasius, Gedas and Wang, Heng and Torresani, Lorenzo},
  booktitle={Icml},
  volume={2},
  number={3},
  pages={4},
  year={2021}
}

@inproceedings{arnab2021vivit,
  title={Vivit: A video vision transformer},
  author={Arnab, Anurag and Dehghani, Mostafa and Heigold, Georg and Sun, Chen and Lu{\v{c}}i{\'c}, Mario and Schmid, Cordelia},
  booktitle={Proceedings of the IEEE/CVF international conference on computer vision},
  pages={6836--6846},
  year={2021}
}

@inproceedings{liu2022video,
  title={Video swin transformer},
  author={Liu, Ze and Ning, Jia and Cao, Yue and Wei, Yixuan and Zhang, Zheng and Lin, Stephen and Hu, Han},
  booktitle={Proceedings of the IEEE/CVF conference on computer vision and pattern recognition},
  pages={3202--3211},
  year={2022}
}

@article{tong2022videomae,
  title={Videomae: Masked autoencoders are data-efficient learners for self-supervised video pre-training},
  author={Tong, Zhan and Song, Yibing and Wang, Jue and Wang, Limin},
  journal={Advances in neural information processing systems},
  volume={35},
  pages={10078--10093},
  year={2022}
}

@inproceedings{wei2022masked,
  title={Masked feature prediction for self-supervised visual pre-training},
  author={Wei, Chen and Fan, Haoqi and Xie, Saining and Wu, Chao-Yuan and Yuille, Alan and Feichtenhofer, Christoph},
  booktitle={Proceedings of the IEEE/CVF conference on computer vision and pattern recognition},
  pages={14668--14678},
  year={2022}
}

@article{akbari2021vatt,
  title={Vatt: Transformers for multimodal self-supervised learning from raw video, audio and text},
  author={Akbari, Hassan and Yuan, Liangzhe and Qian, Rui and Chuang, Wei-Hong and Chang, Shih-Fu and Cui, Yin and Gong, Boqing},
  journal={Advances in neural information processing systems},
  volume={34},
  pages={24206--24221},
  year={2021}
}

@article{wang2022internvideo,
  title={Internvideo: General video foundation models via generative and discriminative learning},
  author={Wang, Yi and Li, Kunchang and Li, Yizhuo and He, Yinan and Huang, Bingkun and Zhao, Zhiyu and Zhang, Hongjie and Xu, Jilan and Liu, Yi and Wang, Zun and others},
  journal={arXiv preprint arXiv:2212.03191},
  year={2022}
}

@inproceedings{venugopalan2015sequence,
  title={Sequence to sequence-video to text},
  author={Venugopalan, Subhashini and Rohrbach, Marcus and Donahue, Jeffrey and Mooney, Raymond and Darrell, Trevor and Saenko, Kate},
  booktitle={Proceedings of the IEEE international conference on computer vision},
  pages={4534--4542},
  year={2015}
}

@inproceedings{sun2019videobert,
  title={Videobert: A joint model for video and language representation learning},
  author={Sun, Chen and Myers, Austin and Vondrick, Carl and Murphy, Kevin and Schmid, Cordelia},
  booktitle={Proceedings of the IEEE/CVF international conference on computer vision},
  pages={7464--7473},
  year={2019}
}

@inproceedings{zhu2020actbert,
  title={Actbert: Learning global-local video-text representations},
  author={Zhu, Linchao and Yang, Yi},
  booktitle={Proceedings of the IEEE/CVF conference on computer vision and pattern recognition},
  pages={8746--8755},
  year={2020}
}

@inproceedings{li2020hero,
  title={Hero: Hierarchical encoder for video+ language omni-representation pre-training},
  author={Li, Linjie and Chen, Yen-Chun and Cheng, Yu and Gan, Zhe and Yu, Licheng and Liu, Jingjing},
  booktitle={Proceedings of the 2020 conference on empirical methods in natural language processing (EMNLP)},
  pages={2046--2065},
  year={2020}
}

@article{luo2020univl,
  title={Univl: A unified video and language pre-training model for multimodal understanding and generation},
  author={Luo, Huaishao and Ji, Lei and Shi, Botian and Huang, Haoyang and Duan, Nan and Li, Tianrui and Li, Jason and Bharti, Taroon and Zhou, Ming},
  journal={arXiv preprint arXiv:2002.06353},
  year={2020}
}

@article{fu2021violet,
  title={Violet: End-to-end video-language transformers with masked visual-token modeling},
  author={Fu, Tsu-Jui and Li, Linjie and Gan, Zhe and Lin, Kevin and Wang, William Yang and Wang, Lijuan and Liu, Zicheng},
  journal={arXiv preprint arXiv:2111.12681},
  year={2021}
}

@inproceedings{li2022align,
  title={Align and prompt: Video-and-language pre-training with entity prompts},
  author={Li, Dongxu and Li, Junnan and Li, Hongdong and Niebles, Juan Carlos and Hoi, Steven CH},
  booktitle={Proceedings of the IEEE/CVF conference on computer vision and pattern recognition},
  pages={4953--4963},
  year={2022}
}

@inproceedings{xu2021videoclip,
  title={Videoclip: Contrastive pre-training for zero-shot video-text understanding},
  author={Xu, Hu and Ghosh, Gargi and Huang, Po-Yao and Okhonko, Dmytro and Aghajanyan, Armen and Metze, Florian and Zettlemoyer, Luke and Feichtenhofer, Christoph},
  booktitle={Proceedings of the 2021 conference on empirical methods in natural language processing},
  pages={6787--6800},
  year={2021}
}

@inproceedings{miech2019howto100m,
  title={Howto100m: Learning a text-video embedding by watching hundred million narrated video clips},
  author={Miech, Antoine and Zhukov, Dimitri and Alayrac, Jean-Baptiste and Tapaswi, Makarand and Laptev, Ivan and Sivic, Josef},
  booktitle={Proceedings of the IEEE/CVF international conference on computer vision},
  pages={2630--2640},
  year={2019}
}

@inproceedings{bain2021frozen,
  title={Frozen in time: A joint video and image encoder for end-to-end retrieval},
  author={Bain, Max and Nagrani, Arsha and Varol, G{\"u}l and Zisserman, Andrew},
  booktitle={Proceedings of the IEEE/CVF international conference on computer vision},
  pages={1728--1738},
  year={2021}
}

@inproceedings{radford2021learning,
  title={Learning transferable visual models from natural language supervision},
  author={Radford, Alec and Kim, Jong Wook and Hallacy, Chris and Ramesh, Aditya and Goh, Gabriel and Agarwal, Sandhini and Sastry, Girish and Askell, Amanda and Mishkin, Pamela and Clark, Jack and others},
  booktitle={International conference on machine learning},
  pages={8748--8763},
  year={2021},
  organization={PmLR}
}

@article{luo2021clip4clip,
  title={Clip4clip: An empirical study of clip for end to end video clip retrieval},
  author={Luo, Huaishao and Ji, Lei and Zhong, Ming and Chen, Yang and Lei, Wen and Duan, Nan and Li, Tianrui},
  journal={arXiv preprint arXiv:2104.08860},
  year={2021}
}

@article{luo2022clip4clip,
  title={Clip4clip: An empirical study of clip for end to end video clip retrieval and captioning},
  author={Luo, Huaishao and Ji, Lei and Zhong, Ming and Chen, Yang and Lei, Wen and Duan, Nan and Li, Tianrui},
  journal={Neurocomputing},
  volume={508},
  pages={293--304},
  year={2022},
  publisher={Elsevier}
}

@inproceedings{ma2022x,
  title={X-clip: End-to-end multi-grained contrastive learning for video-text retrieval},
  author={Ma, Yiwei and Xu, Guohai and Sun, Xiaoshuai and Yan, Ming and Zhang, Ji and Ji, Rongrong},
  booktitle={Proceedings of the 30th ACM international conference on multimedia},
  pages={638--647},
  year={2022}
}

@article{fang2021clip2video,
  title={Clip2video: Mastering video-text retrieval via image clip},
  author={Fang, Han and Xiong, Pengfei and Xu, Luhui and Chen, Yu},
  journal={arXiv preprint arXiv:2106.11097},
  year={2021}
}

@article{alayrac2022flamingo,
  title={Flamingo: a visual language model for few-shot learning},
  author={Alayrac, Jean-Baptiste and Donahue, Jeff and Luc, Pauline and Miech, Antoine and Barr, Iain and Hasson, Yana and Lenc, Karel and Mensch, Arthur and Millican, Katherine and Reynolds, Malcolm and others},
  journal={Advances in neural information processing systems},
  volume={35},
  pages={23716--23736},
  year={2022}
}

@inproceedings{li2023blip,
  title={Blip-2: Bootstrapping language-image pre-training with frozen image encoders and large language models},
  author={Li, Junnan and Li, Dongxu and Savarese, Silvio and Hoi, Steven},
  booktitle={International conference on machine learning},
  pages={19730--19742},
  year={2023},
  organization={PMLR}
}

@inproceedings{zhang2023video,
  title={Video-llama: An instruction-tuned audio-visual language model for video understanding},
  author={Zhang, Hang and Li, Xin and Bing, Lidong},
  booktitle={Proceedings of the 2023 conference on empirical methods in natural language processing: system demonstrations},
  pages={543--553},
  year={2023}
}

@inproceedings{maaz2024video,
  title={Video-chatgpt: Towards detailed video understanding via large vision and language models},
  author={Maaz, Muhammad and Rasheed, Hanoona and Khan, Salman and Khan, Fahad},
  booktitle={Proceedings of the 62nd Annual Meeting of the Association for Computational Linguistics (Volume 1: Long Papers)},
  pages={12585--12602},
  year={2024}
}

@inproceedings{li2024llama,
  title={Llama-vid: An image is worth 2 tokens in large language models},
  author={Li, Yanwei and Wang, Chengyao and Jia, Jiaya},
  booktitle={European Conference on Computer Vision},
  pages={323--340},
  year={2024},
  organization={Springer}
}

@inproceedings{ren2024timechat,
  title={Timechat: A time-sensitive multimodal large language model for long video understanding},
  author={Ren, Shuhuai and Yao, Linli and Li, Shicheng and Sun, Xu and Hou, Lu},
  booktitle={Proceedings of the IEEE/CVF Conference on Computer Vision and Pattern Recognition},
  pages={14313--14323},
  year={2024}
}

@article{wang2023internvid,
  title={Internvid: A large-scale video-text dataset for multimodal understanding and generation},
  author={Wang, Yi and He, Yinan and Li, Yizhuo and Li, Kunchang and Yu, Jiashuo and Ma, Xin and Li, Xinhao and Chen, Guo and Chen, Xinyuan and Wang, Yaohui and others},
  journal={arXiv preprint arXiv:2307.06942},
  year={2023}
}

@article{nan2024openvid,
  title={Openvid-1m: A large-scale high-quality dataset for text-to-video generation},
  author={Nan, Kepan and Xie, Rui and Zhou, Penghao and Fan, Tiehan and Yang, Zhenheng and Chen, Zhijie and Li, Xiang and Yang, Jian and Tai, Ying},
  journal={arXiv preprint arXiv:2407.02371},
  year={2024}
}

@article{chen2024sharegpt4video,
  title={Sharegpt4video: Improving video understanding and generation with better captions},
  author={Chen, Lin and Wei, Xilin and Li, Jinsong and Dong, Xiaoyi and Zhang, Pan and Zang, Yuhang and Chen, Zehui and Duan, Haodong and Lin, Bin and Tang, Zhenyu and others},
  journal={Advances in Neural Information Processing Systems},
  volume={37},
  pages={19472--19495},
  year={2024}
}

@misc{Farre2024FineVideo,
  title={FineVideo},
  author={Farré, Miquel and Marafioti, Andi and Tunstall, Lewis and Von Werra, Leandro and Wolf, Thomas},
  year={2024},
  howpublished={\url{https://huggingface.co/datasets/HuggingFaceFV/finevideo}},
}

@article{li2025videochat,
  title={Videochat: Chat-centric video understanding},
  author={Li, KunChang and He, Yinan and Wang, Yi and Li, Yizhuo and Wang, Wenhai and Luo, Ping and Wang, Yali and Wang, Limin and Qiao, Yu},
  journal={Science China Information Sciences},
  volume={68},
  number={10},
  pages={200102},
  year={2025},
  publisher={Springer}
}

@article{luo2023valley,
  title={Valley: Video assistant with large language model enhanced ability},
  author={Luo, Ruipu and Zhao, Ziwang and Yang, Min and Yang, Zheming and Qiu, Minghui and Wei, Zhongyu and Wang, Yanhao and Chen, Cen},
  journal={ACM Transactions on Multimedia Computing, Communications and Applications},
  year={2023},
  publisher={ACM New York, NY}
}

@article{zhang2024llava,
  title={Llava-video: Video instruction tuning with synthetic data},
  author={Zhang, Yuanhan and Wu, Jinming and Li, Wei and Li, Bo and Ma, Zejun and Liu, Ziwei and Li, Chunyuan},
  journal={arXiv preprint arXiv:2410.02713},
  year={2024}
}

@inproceedings{krishna2017dense,
  title={Dense-captioning events in videos},
  author={Krishna, Ranjay and Hata, Kenji and Ren, Frederic and Fei-Fei, Li and Carlos Niebles, Juan},
  booktitle={Proceedings of the IEEE international conference on computer vision},
  pages={706--715},
  year={2017}
}

@inproceedings{gao2017tall,
  title={Tall: Temporal activity localization via language query},
  author={Gao, Jiyang and Sun, Chen and Yang, Zhenheng and Nevatia, Ram},
  booktitle={Proceedings of the IEEE international conference on computer vision},
  pages={5267--5275},
  year={2017}
}

@inproceedings{anne2017localizing,
  title={Localizing moments in video with natural language},
  author={Anne Hendricks, Lisa and Wang, Oliver and Shechtman, Eli and Sivic, Josef and Darrell, Trevor and Russell, Bryan},
  booktitle={Proceedings of the IEEE international conference on computer vision},
  pages={5803--5812},
  year={2017}
}

@article{lei2021detecting,
  title={Detecting moments and highlights in videos via natural language queries},
  author={Lei, Jie and Berg, Tamara L and Bansal, Mohit},
  journal={Advances in Neural Information Processing Systems},
  volume={34},
  pages={11846--11858},
  year={2021}
}

@inproceedings{soldan2022mad,
  title={Mad: A scalable dataset for language grounding in videos from movie audio descriptions},
  author={Soldan, Mattia and Pardo, Alejandro and Alc{\'a}zar, Juan Le{\'o}n and Caba, Fabian and Zhao, Chen and Giancola, Silvio and Ghanem, Bernard},
  booktitle={Proceedings of the IEEE/CVF Conference on Computer Vision and Pattern Recognition},
  pages={5026--5035},
  year={2022}
}

@inproceedings{grauman2022ego4d,
  title={Ego4d: Around the world in 3,000 hours of egocentric video},
  author={Grauman, Kristen and Westbury, Andrew and Byrne, Eugene and Chavis, Zachary and Furnari, Antonino and Girdhar, Rohit and Hamburger, Jackson and Jiang, Hao and Liu, Miao and Liu, Xingyu and others},
  booktitle={Proceedings of the IEEE/CVF conference on computer vision and pattern recognition},
  pages={18995--19012},
  year={2022}
}

@article{yang2023vidchapters,
  title={Vidchapters-7m: Video chapters at scale},
  author={Yang, Antoine and Nagrani, Arsha and Laptev, Ivan and Sivic, Josef and Schmid, Cordelia},
  journal={Advances in Neural Information Processing Systems},
  volume={36},
  pages={49428--49444},
  year={2023}
}

@inproceedings{yu2019activitynet,
  title={Activitynet-qa: A dataset for understanding complex web videos via question answering},
  author={Yu, Zhou and Xu, Dejing and Yu, Jun and Yu, Ting and Zhao, Zhou and Zhuang, Yueting and Tao, Dacheng},
  booktitle={Proceedings of the AAAI conference on artificial intelligence},
  volume={33},
  number={01},
  pages={9127--9134},
  year={2019}
}

@inproceedings{xiao2021next,
  title={Next-qa: Next phase of question-answering to explaining temporal actions},
  author={Xiao, Junbin and Shang, Xindi and Yao, Angela and Chua, Tat-Seng},
  booktitle={Proceedings of the IEEE/CVF conference on computer vision and pattern recognition},
  pages={9777--9786},
  year={2021}
}

@inproceedings{li2024mvbench,
  title={Mvbench: A comprehensive multi-modal video understanding benchmark},
  author={Li, Kunchang and Wang, Yali and He, Yinan and Li, Yizhuo and Wang, Yi and Liu, Yi and Wang, Zun and Xu, Jilan and Chen, Guo and Luo, Ping and others},
  booktitle={Proceedings of the IEEE/CVF Conference on Computer Vision and Pattern Recognition},
  pages={22195--22206},
  year={2024}
}

@article{wu2024longvideobench,
  title={Longvideobench: A benchmark for long-context interleaved video-language understanding},
  author={Wu, Haoning and Li, Dongxu and Chen, Bei and Li, Junnan},
  journal={Advances in Neural Information Processing Systems},
  volume={37},
  pages={28828--28857},
  year={2024}
}

@inproceedings{fu2025video,
  title={Video-mme: The first-ever comprehensive evaluation benchmark of multi-modal llms in video analysis},
  author={Fu, Chaoyou and Dai, Yuhan and Luo, Yongdong and Li, Lei and Ren, Shuhuai and Zhang, Renrui and Wang, Zihan and Zhou, Chenyu and Shen, Yunhang and Zhang, Mengdan and others},
  booktitle={Proceedings of the IEEE/CVF conference on computer vision and pattern recognition},
  pages={24108--24118},
  year={2025}
}

@inproceedings{wang2025lvbench,
  title={Lvbench: An extreme long video understanding benchmark},
  author={Wang, Weihan and He, Zehai and Hong, Wenyi and Cheng, Yean and Zhang, Xiaohan and Qi, Ji and Ding, Ming and Gu, Xiaotao and Huang, Shiyu and Xu, Bin and others},
  booktitle={Proceedings of the IEEE/CVF International Conference on Computer Vision},
  pages={22958--22967},
  year={2025}
}

@inproceedings{soucek2024transnet,
  title={Transnet v2: An effective deep network architecture for fast shot transition detection},
  author={Soucek, Tom{\'a}s and Lokoc, Jakub},
  booktitle={Proceedings of the 32nd ACM International Conference on Multimedia},
  pages={11218--11221},
  year={2024}
}

@misc{faster_whisper,
  title = {faster-whisper},
  author = {SYSTRAN},
  howpublished = {\url{https://github.com/SYSTRAN/faster-whisper}},
  year = {2024}
}

@inproceedings{radford2023robust,
  title={Robust speech recognition via large-scale weak supervision},
  author={Radford, Alec and Kim, Jong Wook and Xu, Tao and Brockman, Greg and McLeavey, Christine and Sutskever, Ilya},
  booktitle={International conference on machine learning},
  pages={28492--28518},
  year={2023},
  organization={PMLR}
}

@article{chu2024qwen2,
  title={Qwen2-audio technical report},
  author={Chu, Yunfei and Xu, Jin and Yang, Qian and Wei, Haojie and Wei, Xipin and Guo, Zhifang and Leng, Yichong and Lv, Yuanjun and He, Jinzheng and Lin, Junyang and others},
  journal={arXiv preprint arXiv:2407.10759},
  year={2024}
}

@inproceedings{kwon2023efficient,
  title={Efficient memory management for large language model serving with pagedattention},
  author={Kwon, Woosuk and Li, Zhuohan and Zhuang, Siyuan and Sheng, Ying and Zheng, Lianmin and Yu, Cody Hao and Gonzalez, Joseph and Zhang, Hao and Stoica, Ion},
  booktitle={Proceedings of the 29th symposium on operating systems principles},
  pages={611--626},
  year={2023}
}

@inproceedings{
hu2022lora,
title={Lo{RA}: Low-Rank Adaptation of Large Language Models},
author={Edward J Hu and yelong shen and Phillip Wallis and Zeyuan Allen-Zhu and Yuanzhi Li and Shean Wang and Lu Wang and Weizhu Chen},
booktitle={International Conference on Learning Representations},
year={2022},
url={https://openreview.net/forum?id=nZeVKeeFYf9}
}

@inproceedings{patraucean2023perception,
  title     = {Perception Test: A Diagnostic Benchmark for Multimodal Video Models},
  author    = {Patraucean, Viorica and
               Smaira, Lucas and
               Gupta, Ankush and
               Recasens, Adria and
               Markeeva, Larisa and
               Banarse, Dylan and
               Koppula, Skanda and
               Heyward, Joseph and
               Malinowski, Mateusz and
               Yang, Yi and
               Doersch, Carl and
               Matejovicova, Tatiana and
               Sulsky, Yury and
               Miech, Antoine and
               Frechette, Alexandre and
               Klimczak, Hanna and
               Koster, Raphael and
               Zhang, Junlin and
               Winkler, Stephanie and
               Aytar, Yusuf and
               Osindero, Simon and
               Damen, Dima and
               Zisserman, Andrew and
               Carreira, Joao},
  booktitle = {Advances in Neural Information Processing Systems},
  year      = {2023}
}

@inproceedings{wu2021star_situated_reasoning,
author={Wu, Bo and Yu, Shoubin and Chen, Zhenfang and Tenenbaum, Joshua B and Gan, Chuang},
title = {{STAR}: A Benchmark for Situated Reasoning in Real-World Videos},
booktitle = {Thirty-fifth Conference on Neural Information Processing Systems (NeurIPS)},
year = {2021}
}

@inproceedings{CLEVRER2020ICLR,
  author    = {Kexin Yi and
               Chuang Gan and
               Yunzhu Li and
               Pushmeet Kohli and
               Jiajun Wu and
               Antonio Torralba and
               Joshua B. Tenenbaum},
  title     = {{CLEVRER:} Collision Events for Video Representation and Reasoning},
  booktitle = {ICLR},
  year      = {2020}
}

@inproceedings{jia2022egotaskqa,
    title = {EgoTaskQA: Understanding Human Tasks in Egocentric Videos},
    author = {Jia, Baoxiong and Lei, Ting and Zhu, Song-Chun and Huang, Siyuan},
    booktitle = {The 36th Conference on Neural Information Processing Systems (NeurIPS 2022) Track on Datasets and Benchmarks},
    year = {2022}
}

@article{jang2019video,
  title={Video question answering with spatio-temporal reasoning},
  author={Jang, Yunseok and Song, Yale and Kim, Chris Dongjoo and Yu, Youngjae and Kim, Youngjin and Kim, Gunhee},
  journal={International Journal of Computer Vision},
  volume={127},
  number={10},
  pages={1385--1412},
  year={2019},
  publisher={Springer}
}

@article{comanici2025gemini,
  title={Gemini 2.5: Pushing the frontier with advanced reasoning, multimodality, long context, and next generation agentic capabilities},
  author={Comanici, Gheorghe and Bieber, Eric and Schaekermann, Mike and Pasupat, Ice and Sachdeva, Noveen and Dhillon, Inderjit and Blistein, Marcel and Ram, Ori and Zhang, Dan and Rosen, Evan and others},
  journal={arXiv preprint arXiv:2507.06261},
  year={2025}
}

@article{Intelligence2024,
 author = {Amazon Artificial General Intelligence},
 title = {The Amazon Nova family of models: Technical report and model card},
 year = {2024},
 url = {https://www.amazon.science/publications/the-amazon-nova-family-of-models-technical-report-and-model-card},
 journal = {Amazon Technical Reports},
}

@misc{bytedanceseed2026seed2,
  title        = {{Seed2.0}},
  author       = {{ByteDance Seed}},
  year         = {2026},
  howpublished = {\url{https://seed.bytedance.com/en/seed2}},
  note         = {Accessed: 2026-05-07}
}

@article{achiam2023gpt,
  title={Gpt-4 technical report},
  author={Achiam, Josh and Adler, Steven and Agarwal, Sandhini and Ahmad, Lama and Akkaya, Ilge and Aleman, Florencia Leoni and Almeida, Diogo and Altenschmidt, Janko and Altman, Sam and Anadkat, Shyamal and others},
  journal={arXiv preprint arXiv:2303.08774},
  year={2023}
}

@inproceedings{chen2024internvl,
  title={Internvl: Scaling up vision foundation models and aligning for generic visual-linguistic tasks},
  author={Chen, Zhe and Wu, Jiannan and Wang, Wenhai and Su, Weijie and Chen, Guo and Xing, Sen and Zhong, Muyan and Zhang, Qinglong and Zhu, Xizhou and Lu, Lewei and others},
  booktitle={CVPR},
  year={2024}
}

@article{hong2025glm,
  title={Glm-4.1 v-thinking: Towards versatile multimodal reasoning with scalable reinforcement learning},
  author={Hong, Wenyi and Yu, Wenmeng and Gu, Xiaotao and Wang, Guo and Gan, Guobing and Tang, Haomiao and Cheng, Jiale and Qi, Ji and Ji, Junhui and Pan, Lihang and others},
  pages={arXiv--2507},
  year={2025}
}

@article{wang2025internvl3,
    title={InternVL3. 5: Advancing Open-Source Multimodal Models in Versatility, Reasoning, and Efficiency},
    author={Wang, Weiyun and Gao, Zhangwei and Gu, Lixin and Pu, Hengjun and Cui, Long and Wei, Xingguang and Liu, Zhaoyang and Jing, Linglin and Ye, Shenglong and Shao, Jie and others},
    journal={arXiv preprint arXiv:2508.18265},
    year={2025}
}

@misc{coreteam2025mimounlockingreasoningpotential,
      title={MiMo: Unlocking the Reasoning Potential of Language Model -- From Pretraining to Posttraining}, 
      author={LLM-Core-Team Xiaomi},
      year={2025},
      eprint={2505.07608},
      archivePrefix={arXiv},
      primaryClass={cs.CL},
      url={https://arxiv.org/abs/2505.07608}, 
}

@article{Qwen-VL,
  title={Qwen-VL: A Versatile Vision-Language Model for Understanding, Localization, Text Reading, and Beyond},
  author={Bai, Jinze and Bai, Shuai and Yang, Shusheng and Wang, Shijie and Tan, Sinan and Wang, Peng and Lin, Junyang and Zhou, Chang and Zhou, Jingren},
  journal={arXiv preprint arXiv:2308.12966},
  year={2023}
}

@misc{googledeepmind2026gemma4,
  title        = {{Gemma 4 Model Card}},
  author       = {{Google DeepMind}},
  year         = {2026},
  howpublished = {\url{https://ai.google.dev/gemma/docs/core/model_card_4}},
  note         = {Accessed: 2026-05-07}
}

@article{hong2024cogvlm2,
  title={CogVLM2: Visual Language Models for Image and Video Understanding},
  author={Hong, Wenyi and Wang, Weihan and Ding, Ming and Yu, Wenmeng and Lv, Qingsong and Wang, Yan and Cheng, Yean and Huang, Shiyu and Ji, Junhui and Xue, Zhao and others},
  journal={arXiv preprint arXiv:2408.16500},
  year={2024}
}

@misc{step3system,
      title={Step-3 is Large yet Affordable: Model-system Co-design for Cost-effective Decoding}, 
      author={StepFun Team},
      year={2025},
      eprint={2507.19427},
      archivePrefix={arXiv},
      primaryClass={cs.LG},
      url={https://arxiv.org/abs/2507.19427}, 
}

@article{li2024llava,
  title={Llava-onevision: Easy visual task transfer},
  author={Li, Bo and Zhang, Yuanhan and Guo, Dong and Zhang, Renrui and Li, Feng and Zhang, Hao and Zhang, Kaichen and Zhang, Peiyuan and Li, Yanwei and Liu, Ziwei and others},
  journal={arXiv preprint arXiv:2408.03326},
  year={2024}
}

@misc{chen2024cgbenchcluegroundedquestionanswering,
      title={CG-Bench: Clue-grounded Question Answering Benchmark for Long Video Understanding}, 
      author={Guo Chen and Yicheng Liu and Yifei Huang and Yuping He and Baoqi Pei and Jilan Xu and Yali Wang and Tong Lu and Limin Wang},
      year={2024},
      eprint={2412.12075},
      archivePrefix={arXiv},
      primaryClass={cs.CV},
      url={https://arxiv.org/abs/2412.12075}, 
}

@inproceedings{Qin_2025, series={SIGIR ’25},
   title={Question-Answering Dense Video Events},
   url={http://dx.doi.org/10.1145/3726302.3729945},
   DOI={10.1145/3726302.3729945},
   booktitle={Proceedings of the 48th International ACM SIGIR Conference on Research and Development in Information Retrieval},
   publisher={ACM},
   author={Qin, Hangyu and Xiao, Junbin and Yao, Angela},
   year={2025},
   month=July, pages={884–894},
   collection={SIGIR ’25} }

@article{mangalam2023egoschema,
  title={Egoschema: A diagnostic benchmark for very long-form video language understanding},
  author={Mangalam, Karttikeya and Akshulakov, Raiymbek and Malik, Jitendra},
  journal={NeurIPS},
  year={2023}
}

@misc{cheng2025videoholmesmllmthinklike,
      title={Video-Holmes: Can MLLM Think Like Holmes for Complex Video Reasoning?}, 
      author={Junhao Cheng and Yuying Ge and Teng Wang and Yixiao Ge and Jing Liao and Ying Shan},
      year={2025},
      eprint={2505.21374},
      archivePrefix={arXiv},
      primaryClass={cs.CV},
      url={https://arxiv.org/abs/2505.21374}, 
}

@misc{chandrasegaran2024hourvideo1hourvideolanguageunderstanding,
      title={HourVideo: 1-Hour Video-Language Understanding}, 
      author={Keshigeyan Chandrasegaran and Agrim Gupta and Lea M. Hadzic and Taran Kota and Jimming He and Cristóbal Eyzaguirre and Zane Durante and Manling Li and Jiajun Wu and Li Fei-Fei},
      year={2024},
      eprint={2411.04998},
      archivePrefix={arXiv},
      primaryClass={cs.CV},
      url={https://arxiv.org/abs/2411.04998}, 
}

@misc{geminiteam2026gemini31pro,
  title        = {{Gemini 3.1 Pro}: A Smarter Model for Your Most Complex Tasks},
  author       = {{The Gemini Team}},
  year         = {2026},
  month        = feb,
  day          = {19},
  howpublished = {\url{https://blog.google/innovation-and-ai/models-and-research/gemini-models/gemini-3-1-pro/}},
  note         = {Accessed: 2026-05-07}
}

@misc{openai2026gpt55,
  title        = {{GPT-5.5} System Card},
  author       = {{OpenAI}},
  year         = {2026},
  howpublished = {\url{https://openai.com/index/gpt-5-5-system-card/}},
  note         = {Accessed: 2026-05-07}
}

@misc{openai2026gpt4omini,
  title        = {{GPT‑4o mini}: advancing cost-efficient intelligence},
  author       = {{OpenAI}},
  year         = {2026},
  howpublished = {\url{https://openai.com/index/gpt-4o-mini-advancing-cost-efficient-intelligence/}},
  note         = {Accessed: 2026-05-07}
}

@article{bai2025qwen3,
  title={Qwen3-vl technical report},
  author={Bai, Shuai and Cai, Yuxuan and Chen, Ruizhe and Chen, Keqin and Chen, Xionghui and Cheng, Zesen and Deng, Lianghao and Ding, Wei and Gao, Chang and Ge, Chunjiang and others},
  journal={arXiv preprint arXiv:2511.21631},
  year={2025}
}

@article{zhu2025internvl3,
  title={Internvl3: Exploring advanced training and test-time recipes for open-source multimodal models},
  author={Zhu, Jinguo and Wang, Weiyun and Chen, Zhe and Liu, Zhaoyang and Ye, Shenglong and Gu, Lixin and Tian, Hao and Duan, Yuchen and Su, Weijie and Shao, Jie and others},
  journal={arXiv preprint arXiv:2504.10479},
  year={2025}
}

@misc{glm2024chatglm,
      title={ChatGLM: A Family of Large Language Models from GLM-130B to GLM-4 All Tools}, 
      author={Team GLM and Aohan Zeng and Bin Xu and Bowen Wang and Chenhui Zhang and Da Yin and Diego Rojas and Guanyu Feng and Hanlin Zhao and Hanyu Lai and Hao Yu and Hongning Wang and Jiadai Sun and Jiajie Zhang and Jiale Cheng and Jiayi Gui and Jie Tang and Jing Zhang and Juanzi Li and Lei Zhao and Lindong Wu and Lucen Zhong and Mingdao Liu and Minlie Huang and Peng Zhang and Qinkai Zheng and Rui Lu and Shuaiqi Duan and Shudan Zhang and Shulin Cao and Shuxun Yang and Weng Lam Tam and Wenyi Zhao and Xiao Liu and Xiao Xia and Xiaohan Zhang and Xiaotao Gu and Xin Lv and Xinghan Liu and Xinyi Liu and Xinyue Yang and Xixuan Song and Xunkai Zhang and Yifan An and Yifan Xu and Yilin Niu and Yuantao Yang and Yueyan Li and Yushi Bai and Yuxiao Dong and Zehan Qi and Zhaoyu Wang and Zhen Yang and Zhengxiao Du and Zhenyu Hou and Zihan Wang},
      year={2024},
      eprint={2406.12793},
      archivePrefix={arXiv},
      primaryClass={id='cs.CL' full_name='Computation and Language' is_active=True alt_name='cmp-lg' in_archive='cs' is_general=False description='Covers natural language processing. Roughly includes material in ACM Subject Class I.2.7. Note that work on artificial languages (programming languages, logics, formal systems) that does not explicitly address natural-language issues broadly construed (natural-language processing, computational linguistics, speech, text retrieval, etc.) is not appropriate for this area.'}
}

@misc{openai2026gpt54,
  title        = {Introducing {GPT‑5.4}},
  author       = {{OpenAI}},
  year         = {2026},
  howpublished = {\url{https://openai.com/index/introducing-gpt-5-4/}},
  note         = {Accessed: 2026-05-07}
}

@misc{anthropic2026claudeopus47,
  title        = {Introducing {Claude Opus 4.7}},
  author       = {{Anthropic}},
  year         = {2026},
  howpublished = {\url{https://www.anthropic.com/news/claude-opus-4-7}},
  note         = {Accessed: 2026-05-07}
}
